%% file: SOT_arxiv.tex
\documentclass[runningheads]{llncs}
\usepackage{graphicx}

\usepackage{tikz}
\usepackage{amsmath,amssymb} 
\usepackage{color}

\usepackage{epsfig}
\usepackage{graphicx}
\usepackage{amsmath}
\usepackage{amssymb}
\usepackage{multirow}
\usepackage{subcaption}
\usepackage[english]{babel}
\usepackage{mathtools}
\usepackage{comment}
\usepackage{arydshln}
\usepackage{xcolor}
\usepackage{pifont}
\usepackage[utf8]{inputenc}
\usepackage[noend]{algpseudocode}
\usepackage{enumitem}
\usepackage[normalem]{ulem}
\DeclareMathOperator*{\argminA}{arg\,min} 
\usepackage{sidecap}

\usepackage{caption}
\usepackage{subcaption}

\usepackage{lmodern} 

\newcommand{\ie}{\textit{i}.\textit{e}.}

\newcommand{\teal}[1]{{\footnotesize \bf\color{teal} #1}}
\newcommand{\pink}[1]{{\footnotesize \bf\color{pink} #1}}
\newcommand{\blue}[1]{{\footnotesize \bf\color{blue} #1}}
\newcommand{\red}[1]{{\footnotesize \bf\color{red} #1}}
\newcommand{\brown}[1]{{\footnotesize \bf\color{olive} #1}}

\usepackage[pagebackref,breaklinks,colorlinks]{hyperref}

\usepackage[capitalize]{cleveref}
\crefname{section}{Sec.}{Secs.}
\Crefname{section}{Section}{Sections}
\Crefname{table}{Table}{Tables}
\crefname{table}{Tab.}{Tabs.}

\usepackage[accsupp]{axessibility}  

\usepackage[width=122mm,left=12mm,paperwidth=146mm,height=193mm,top=12mm,paperheight=217mm]{geometry}

\begin{document}
\pagestyle{headings}
\mainmatter
\def\ECCVSubNumber{6933}  

\title{The Self-Optimal-Transport Feature Transform}


%
\author{Daniel Shalam\and
Simon Korman}
%
%
\institute{University of Haifa, Israel}
\maketitle

\input{abstract.tex} 

\section{Introduction}

\input{introduction.tex} 

\section{Related Work}

\input{related_work.tex}

\section{Method}

\input{method.tex} 

\section{Implementation details}

\input{implementation_details} 

\section{Results}

\input{results}

\section{Conclusions, Limitations and Future Work}

\input{conclusions} 

{\small
\bibliographystyle{ieee_fullname}
\bibliography{reference}
}

\newpage
\appendix
\renewcommand\thesection{\Alph{section}}
\renewcommand\thesubsection{\thesection.\arabic{subsection}}
\noindent{\LARGE{Appendix}}

\input{SOT_appendix}

\end{document}

%% file: abstract.tex
\begin{abstract}

The Self-Optimal-Transport (SOT) feature transform is designed to upgrade the set of features of a data instance to facilitate downstream matching or grouping related tasks. 
The transformed set encodes a rich representation of high order relations between the instance features. Distances  between transformed features capture their \textit{direct} original similarity and their \textit{third party} `agreement' regarding similarity to other features in the set. 
A particular min-cost-max-flow fractional matching problem, whose entropy regularized version can be approximated by an optimal transport (OT) optimization, results in our transductive transform which is efficient, differentiable, equivariant, parameterless and probabilistically interpretable.
Empirically, the transform is highly effective and flexible in its use, consistently improving networks it is inserted into, in a variety of tasks and training schemes. We demonstrate its merits through the problem of unsupervised clustering and its efficiency and wide applicability for few-shot-classification, with state-of-the-art results, and large-scale person re-identification. 

\end{abstract}






%% file: introduction.tex
In this work, we reassess the design and functionality of features for $instance$-$specific$ problems.
In such problems, typically, features computed at test time are mainly compared relative to one another, and less so to the features seen at training time. For such problems the standard practice of learning a generic feature extractor during training and applying it at test time might be sub-optimal. 

We aim at finding training and inference schemes that take into account these considerations, being able to exploit large corpuses of training data to learn features that can easily adapt, or be relevant, to the test time task. Our approach to doing so will be in the form of a \textit{feature transform} that jointly re-embeds the set of features of an instance in a way that resembles how recently popular self-attention mechanisms and Transformers \cite{ramachandran2019stand,lee2019set,mialon2021trainable,khan2021transformers} re-embed sets of features.

Being at the low-to-mid-level of most relevant architectures, advances in such feature re-embeddings have a direct impact and wide applicability in instance-specific problems such as few-shot classification {\cite{ravi2017optimization}}, clustering {\cite{van2020scan}}, patch matching {\cite{korman2015coherency}} and person re-identification {\cite{ye2021deep}}, to name but a few.

\begin{figure*}[t]
\centering\vspace{-2pt}
    \includegraphics[width=0.99\textwidth]{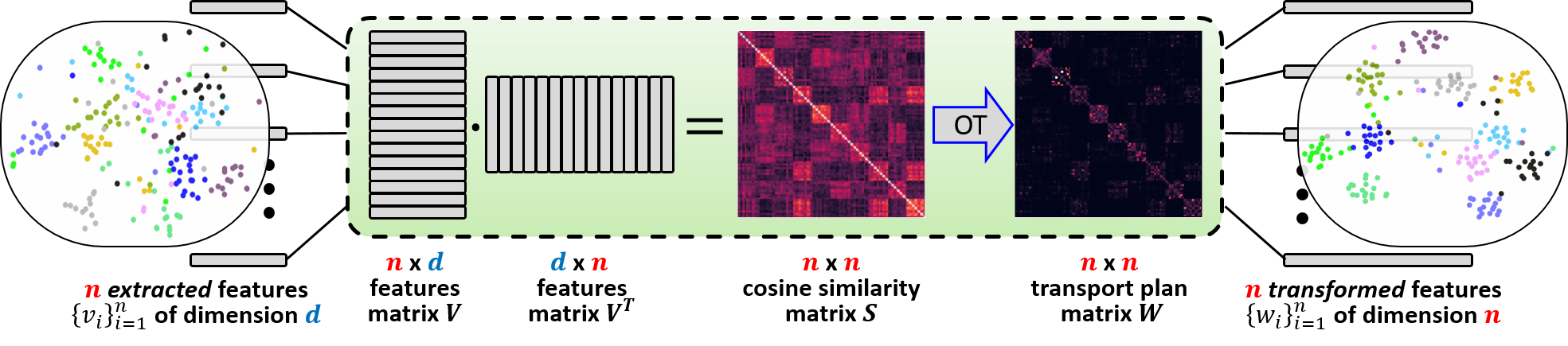} 
   \vspace{-8pt}
   \caption{    {\fontsize{8.5}{8.5} \selectfont
   \textbf{The SOT transform:} 
   Its input is a set of $n$ $d$-dimensional features (each shown as a horizontal gray rectangle, and as a colored point in the input embedding space where color depicts class label or equivalent). Processing is as follows: The unit length (normalized) features are arranged in an $n\times d$ matrix for computing a pairwise $n\times n$ cosine similarity matrix $S$. Then, the transport-plan matrix $W$ (given a specific OT instance that depends on $S$) is computed using several Sinkhorn \cite{cuturi2013sinkhorn} iterations. Finally, the transformed output features are basically the rows of the matrix $W$. As we claim and observe in real results, the features are re-embedded in a way that is consistently superior for downstream grouping and matching tasks (observed the better formation of the embedded points, e.g. towards applying a linear classifier or an off-the-shelf clustering procedure).}
   }
    \label{fig.SOT}\vspace{-2pt}
\end{figure*}

The general idea of the Self-Optimal-Transport (SOT) feature transform that we propose is depicted and explained in Fig. \ref{fig.SOT}, as part of the general design of networks that work on sets
which we illustrate in Fig. \ref{fig.network}.

\vspace{-1pt}
\subsection{Overview}\vspace{-4pt}

We are given an instance of some inference problem, in the form of a set of $n$ items $\{x_i\}_{i=1}^n$, represented as vectors in $\mathbb{R}^D$, for a fixed dimension $D$. 
A generic neural-network (Fig. \ref{fig.network} Left) typically uses a feature embedding (extractor) $F:\mathbb{R}^D\rightarrow\mathbb{R}^d$ (with $d\ll D$), which is applied independently on each input item, to obtain a set of features ${V}=\{v_i\}_{i=1}^n=\{F(x_i)\}_{i=1}^n$.
The features ${V}$ might be of high quality (concise, unique, descriptive), but are limited in representation since they are extracted based on knowledge acquired for similar examples at train time, with no context of the test time instance that they are part of. 

We adapt a rather simple framework (Fig. \ref{fig.network} Right) in which some \textit{transform} acts on the entire set of instance features. The idea is to jointly process the set of features to output an updated set (one for each input feature), that re-embeds each feature in light of the joint statistics of the entire instance. The proposed features transform can be seen as a special case of an attention mechanism \cite{ramachandran2019stand} specialized to features of instance-specific tasks, with required adaptations. Techniques developed here borrow from and might lend to those used in  
    set-to-set \cite{zaheer2017deep,ye2020few,maron2020learning}, 
    self-attention \cite{ramachandran2019stand,mialon2021trainable}
    and transformer \cite{lee2019set,khan2021transformers}
    architectures.

\vspace{-1pt}
\subsection{Contributions}\vspace{-4pt}

\begin{figure*}[t]
\centering \vspace{-2pt} 
\begin{tabular}{c c}
     \includegraphics[height=0.305\textwidth]{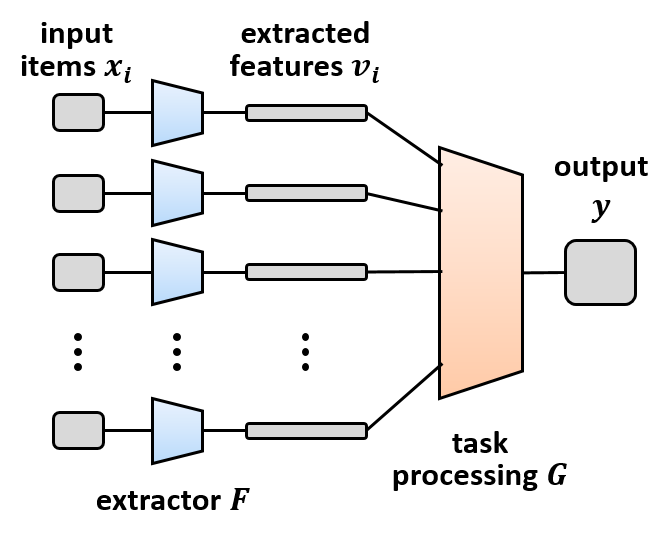} 
     &  \includegraphics[height=0.305\textwidth]{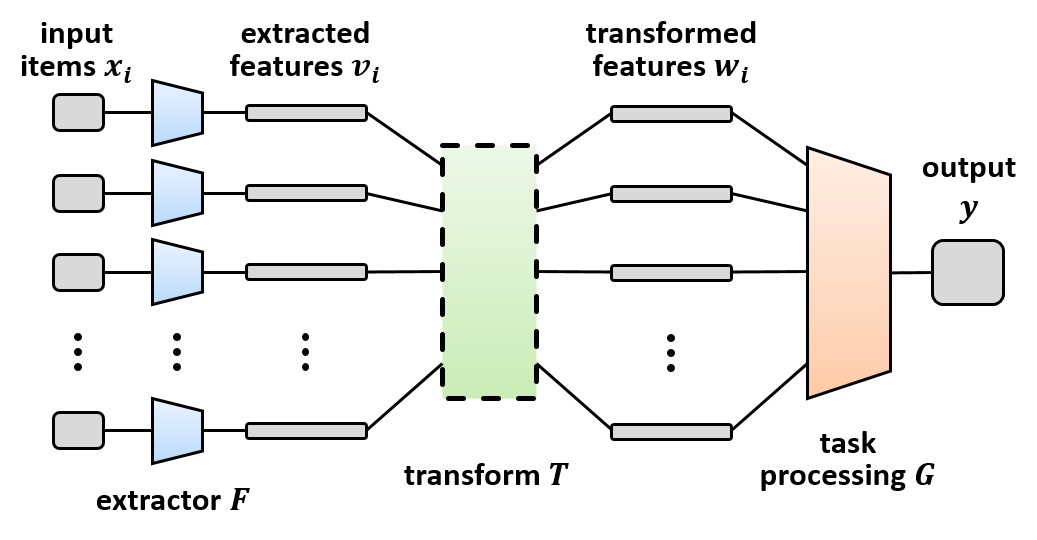} \\
\end{tabular}
   \vspace{-10pt}\caption{    {\fontsize{8.5}{8.5} \selectfont
   \textbf{Generic designs of networks that act on \textit{sets} of inputs.} These cover relevant architectures, e.g. for few-shot-classification and clustering. \textbf{Left:} A generic network for processing a set of input items typically follows the depicted structure: (i) Each item \textit{separately} goes through a common feature extractor $F$. (ii) The set of extracted features is the input to a downstream task processing module $G$. ; \textbf{Right:} A more general structure in which the extracted features undergo a \textit{joint} processing by a transform $T$. Our SOT transform (as well as other attention mechanisms) is of this type and its high-level design (within the `green' module) is detailed in Fig. \ref{fig.SOT}.}}
    \label{fig.network}\vspace{-2pt}
\end{figure*}

We propose a parameter-less transform $T$, which can be used as a drop-in addition that can convert a conventional network to an instance-aware one (e.g. from \cref{fig.network} Left to Right). 
We propose an optimal-transport based feature transform which is shown to have the following attractive set of qualities. 
(i) \textit{efficiency}: having real-time inference; 
(ii) \textit{differentiability}: allowing end-to-end training of the entire `embedding-transform-inference' pipeline of  Fig. \ref{fig.network} Right; 
(iii) \textit{equivariance}: ensuring that the embedding works coherently under any order of the input items; 
(iv) \textit{capturing relative similarity}: 
The comparison of embedded vectors will include both direct and indirect (third-party) similarity information between the input features; 
(v) \textit{probabilistic interpretation}: 
each embedded feature will encode its distribution of similarities to all other features, by conforming to a doubly-stochastic constraint; 
(vi) \textit{instance-aware dimensionality}: embedding dimension (capacity) is adaptive to input size (complexity).

We provide a detailed analysis of our method and show its flexibility and ease of application to a wide variety of tasks, by incorporating it in leading methods of each kind.
A controlled experiment on unsupervised clustering is used to verify its performance, with a detailed analysis. For few-shot-classification we perform an extensive comparison to existing work on several benchmarks, showing that SOT achieves new state-of-art results. 
Finally, we show that SOT is easily applicable to large-scale benchmarks by using the person re-identification task, for which it consistently improves state-of-art networks that it is incorporated into.

%% file: related_work.tex
\subsection{Related Techniques}
\paragraph{Set-to-set or set-to-feature functions}
Our method can clearly be categorized along with recent techniques that act jointly on a set of items (typically features) to output an updated set (or a single feature), which are typically used for downstream inference tasks on the items individually, or as a set. 
The pioneering Deep-Sets~\cite{zaheer2017deep}  formalized fundamental requirements from architectures that process sets. Point-Net~\cite{qi2017pointnet} presented an influential design that learns local and global features on 3D point-clouds, while Maron \textit{et.al.}  \cite{maron2020learning} study layer designs that approximate equivariant and invariant functions. 
Unlike the proposed SOT transform, the joint processing in these methods is very limited, amounting to (Siamese) weight-sharing between separate processes and simple joint aggregations like average pooling. 

\paragraph{Self-Attention}
The introduction of Relational Networks \cite{santoro2017simple} and transformers~\cite{vaswani2017attention} and their initial applications in vision models~\cite{ramachandran2019stand} have lead to a surge of following successful works~\cite{khan2021transformers}, many of which are dedicated to few-shot-learning, such as ReNet \cite{ReNet}, DeepEMD \cite{DeepEMD} and FEAT\cite{ye2020few}. Different from these methods, SOT is parameterless, and hence can work at test-time on any pre-trained network. In addition, SOT is the only method that provides an explicit probabilistic global interpretation of the instance data.

\paragraph{Optimal Transport}
Optimal transport (OT) problems are tightly related to measuring and calculating distances between distributions or sets of features. 
In~\cite{cuturi2013sinkhorn} Cuturi popularized the Sinkhorn algorithm which is a simple, differentiable and fast approximation of entropy-regularized OT problems. 
The Set transformer~\cite{lee2019set} uses an OT-based clustering algorithm, SuperGlue~\cite{sarlin2020superglue} uses OT in an end-to-end manner for feature-point matching, and many state-of-the-art methods in few-shot learning, which we review next, have adopted the Sinkhorn algorithm to model relations between features and class representations. 
The differentiability and efficiency of regularized OT solvers has recently been shown useful in related domains, to derive a differentiable `top-k' operator \cite{xie2020differentiable} or for style transfer applications, by viewing styles as a distributions between which distances are approximated \cite{kolkin2019style}. 
In this work we focus on \textit{self} applications of OT, which enables concise modelings of the relative similarities within a set of items.

\subsection{Few-Shot-Classification (FSC)}
\textit{Few-Shot-Classification}~\cite{vinyals2016matching} is a branch of few-shot-learning in which a classifier needs to learn to recognize classes unseen given a limited number of labeled examples. A FSC task is a self-contained instance that includes both support (labeled) and query (unlabeled) items, hence is a clear instance-specific setup which SOT can handle.

Some leading FSC approaches follow the \textit{meta-learning} (or ``learn-to-learn") principle in which the training data is split into tasks (or episodes) mimicking the test time tasks to which the learner is required to generalize. 
The celebrated MAML~\cite{finn2017model} ``learns to fine-tune" by learning a network initialization from which it can adapt to a novel set of classes with very few gradient update steps on the labeled examples. 
In ProtoNet~\cite{snell2017prototypical}, a learner is meta-trained to predict query feature classes, based on distances from support (labeled) class-prototypes in the embedding space. 
The trainable version of SOT is a meta-learning algorithm, but unlike the above, it is transductive (see ahead) and exploits the task items as a set, while directly assessing the relative similarity relations between its items.

Subsequent works \cite{chen2018closer,dhillon2019baseline} have questioned the benefits of meta-learning, advocating the standard transfer learning procedure of fine-tuning pre-trained networks. In particular, they demonstrate the advantages of using larger and more powerful feature-encoding architectures, as well as the employment of \textit{transductive} inference, which fully exploits the data of the inference task, including unlabeled images.
As mentioned, SOT is a purely transductive method, but it is significantly more flexible in its assumptions, since the transform is based on a general probabilistic grouping action. It does not make any assumptions on (nor does it need to know) the number of classes and the number of items per class in an instance. 

More recently, \textit{attention mechanisms} were shown to be effective for FSC. We have reviewed some relevant works of this line in the previous section. 

Finally, a large number of works have adopted the Sinkhorn Algorithm~\cite{cuturi2013sinkhorn} as a parameterless unsupervised classifier that computes fractional matchings between query embeddings and class centers. 
Many leading FSC works use this approach, including Laplacian-Shot \cite{ziko2020laplacian}, CentroidNet \cite{huang2019few} and PT-MAP \cite{hu2020leveraging}. The current state-of-the-art is set by the recent Sill-Net \cite{zhang2021sill}, which augments training samples with illumination features that are separated from the images in feature space and by PT-MAP-sf \cite{chen2021few}, who propose a DCT-based feature embedding network, encoding detailed frequency-domain information that complements the standard spatial domain features. Both methods are based on PT-MAP \cite{hu2020leveraging}. 
SOT uses Sinkhorn to solve an entirely different OT problem - that of matching the set of features to itself, rather than against class representations. Nevertheless, SOT can be incorporated into these methods, immediately after their feature extraction stage.

\subsection{Unsupervised Clustering and Person Re-Identification (Re-ID)}
These domains are not at the focus of this work therefore we only briefly give some useful pointers for the sake of brevity.

%

Unsupervised image clustering is an active area of research, with standardised evaluation protocols (from Cifar-10~\cite{krizhevsky2009learning} to different subsets of ImageNet~\cite{deng2009imagenet}).
Prominent works in this area include 
Deep Adaptive Clustering (DAC) \cite{chang2017deep}, 
Invariant Information Clustering (IIC) \cite{ji2019invariant} and 
SCAN \cite{van2020scan}.
%
Clustering has recently gained popularity as a means for self-supervision in feature learning, showing excellent results on unsupervised image classification. See for example Deep-Cluster~\cite{caron2018deep} and SWAV~\cite{Caron2020UnsupervisedLO}.
Clustering is a clear case instance specific problem, since most information is relative and unrelated directly to other training data. Our transform can hence be used to upgrade the feature representation quality.

We chose the Re-ID application as another instance-specific problem, which from our point of view differs from the others considered in two main aspects which we find attractive: (i) The tasks are of larger scale - querying thousands of identities against a target set of (tens of) thousands. (ii) The data is much more real-world compared to the carefully curated classification and clustering tasks. 
See \cite{ye2021deep} for an excellent recent and comprehensive survey on the topic. 

%% file: method.tex
Assume we are given a task which consists of an inference problem over a set of $n$ items $\{x_i\}_{i=1}^n$, where each of the items belongs to a space of input items $\Omega\subseteq\mathbb{R}^D$. 
%
The inference task can be modeled as $f_\theta(\{x_i\}_{i=1}^n)$, using a learned function $f_\theta$, which acts on the set of input items and is parameterized by a set of parameters $\theta$.

Typically, such functions combine an initial feature extraction stage that is applied independently to each input item, with a subsequent stage of (separate or joint) processing of the feature vectors (see Fig. \ref{fig.network} Left or Right, respectively).

That is, the function $f_\theta$ takes the form $f_\theta(\{x_i\}_{i=1}^n)=G_{\psi}(\{F_{\phi}(x_i)\}_{i=1}^n)$, where $F_{\phi}$ is the feature extractor (or embedding network) and $G_{\psi}$ is the task inference function, parameterized by $\phi$ and $\psi$ respectively, where $\theta=\phi\cup\psi$. 

The feature embedding $F:\mathbb{R}^D\rightarrow\mathbb{R}^d$, usually in the form of a neural-network (with $d\ll D$), could be either pre-trained, or trained in the context of the task function $f$, along with the inference function $G$.

For an input $\{x_i\}_{i=1}^n$, let us define the set of features $\{v_i\}_{i=1}^n=\{F(x_i)\}_{i=1}^n$. In the following, we consider these sets of input vectors and features as real-valued row-stacked matrices $\mathcal{X}\in \mathbb{R}^{n\times D}$ and $\mathcal{V}\in\mathbb{R}^{n\times d}$.

We suggest a novel re-embedding of the feature set $\mathcal{V}$, using a transform that we denote by $T$, in order to obtain a new set of features $\mathcal{W}=T(\mathcal{V})$, where $\mathcal{W}\in\mathbb{R}^{n\times n}$.
The new feature set $\mathcal{W}$ has an explicit probabilistic interpretation, which is specifically suited for tasks related to classification, matching or grouping of items in the input set $\mathcal{X}$.
In particular, $\mathcal{W}$ will be a symmetric, doubly-stochastic matrix, where the entry $w_{ij}$ (for $i\neq j$) gives the probability that items $x_i$ and $x_j$ belong to the same class or cluster. 

The proposed transform $T:\mathbb{R}^{n\times d}\rightarrow \mathbb{R}^{n\times n}$ (see Fig.~\ref{fig.SOT}) acts on the original feature set $\mathcal{V}$ as follows. It begins by computing the squared Euclidean pairwise distances matrix $\mathcal{D}$, namely, $d_{ij}=||v_i-v_j||^2$, which can be computed efficiently as $d_{ij}=2(1-cos(v_i,v_j))=2(1-v_i\cdot v_j^T)$, assuming that the rows of $\mathcal{V}$ are unit normalized. Or in a compact form, $\mathcal{D}=2(\textbf{1}-\mathcal{S})$, where
$\textbf{1}$ is the all ones $n\times n$ matrix and 
$\mathcal{S}=\mathcal{V}\cdot\mathcal{V}^T$ is the cosine similarity matrix of $\mathcal{V}$. 

$\mathcal{W}$ will be computed as the optimal transport (OT) plan matrix between the $n$-dimensional all-ones vector $\textbf{1}_n$ and itself, under the cost matrix $\mathcal{D}_\infty$, which is the distance matrix $\mathcal{D}$ with a very (infinitely) large scalar replacing each of the entries on its diagonal (which were all zero).
Explicitly, let $\mathcal{D}_\infty=\mathcal{D}+\alpha I$, where $\alpha$ is a very (infinitely) large constant and $I$ is an $n\times n$ identity matrix.

$\mathcal{W}$ is defined to be the doubly-stochastic matrix\footnote{a square ($n\times n$) matrix of non-negative real values, each of whose rows and columns sums to $1$} that is the minimizer of the functional
    \begin{equation} \label{eq.fractional_matching}
      \mathcal{W}=\argminA_{\mathcal{W}\in B_n}\:\langle \mathcal{D}_\infty,\mathcal{W}\rangle
    \end{equation}
    where $B_n$ is the set (known as the Birkhoff polytope) of $n\times n$ doubly-stochastic matrices and $\langle\cdot,\cdot\rangle$ stands for the Frobenius (standard) dot-product.

This objective can be minimized using simplex or interior point methods with complexity $\Theta(n^3\log n)$. In practice, we use the highly efficient Sinkhorn-Knopp method \cite{cuturi2013sinkhorn}, which is an iterative scheme that optimizes an entropy-regularized version of the problem, where each iteration takes $\Theta(n^2)$. Namely:
    \begin{equation} \label{eq.entropy_min}
        \mathcal{W}=\argminA_{\mathcal{W}\in B_n}\:\langle \mathcal{D}_\infty,\mathcal{W}\rangle-\frac{1}{\lambda}h(\mathcal{W})
    \end{equation} \label{eq:sinkhorn_objective}
where $h(\mathcal{W})=-\sum_{i,j} w_{ij} \log(w_{ij})$ is the Shannon entropy of $\mathcal{W}$ and $\lambda$ is the entropy regularization parameter.

The \textit{transport-plan} matrix $\mathcal{W}$ that is the minimizer of \cref{eq.entropy_min} is the result of our transform, i.e. $\mathcal{W}=T(\mathcal{V})$ and each of its rows is the re-embedding of each of the corresponding features (rows) in $\mathcal{V}$. Recall that $\mathcal{W}$ is doubly-stochastic and note that it is symmetric\footnote{The symmetry of $\mathcal{W}$ is as a result of the symmetry of $\mathcal{D}$ and the double-stochasticity of $\mathcal{W}$.}. We next explain its probabilistic interpretation.

The optimization problem in \cref{eq.fractional_matching} can be written more explicitly as follows:
        \begin{equation}
        \begin{aligned}
        \min_{\mathcal{W}} \; \langle \mathcal{D}_\infty,\mathcal{W}\rangle \quad \quad
        \textrm{s.t.} \quad \quad & \mathcal{W}\cdot \textbf{1}_n=\mathcal{W}^T\cdot \textbf{1}_n= \textbf{1}_n
        \end{aligned} \label{eq.opt_D_inf}
        \end{equation}

 which can be seen to be the same as:
        \begin{equation}
        \begin{aligned}
        \min_{\mathcal{W}} \; \langle \mathcal{D},\mathcal{W}\rangle 
        \quad \quad
        \textrm{s.t.} \quad \quad & \mathcal{W}\cdot \textbf{1}_n=\mathcal{W}^T\cdot \textbf{1}_n= \textbf{1}_n \\
        & w_{ii}=0 \quad\text{for}\quad i = 1,\dots n
        \end{aligned} \label{eq.opt_fractional_matching}
        \end{equation}
        since the use of the infinite weights on the diagonal of $\mathcal{D}_\infty$ is equivalent to using the original $\mathcal{D}$ with a constraint of zeros along the diagonal of $\mathcal{W}$.
        
\begin{figure}[]
\centering  
\setlength{\tabcolsep}{0.16em} 
\small
\begin{tabular}{c c c c c c}
     \includegraphics[height=0.085\textheight]{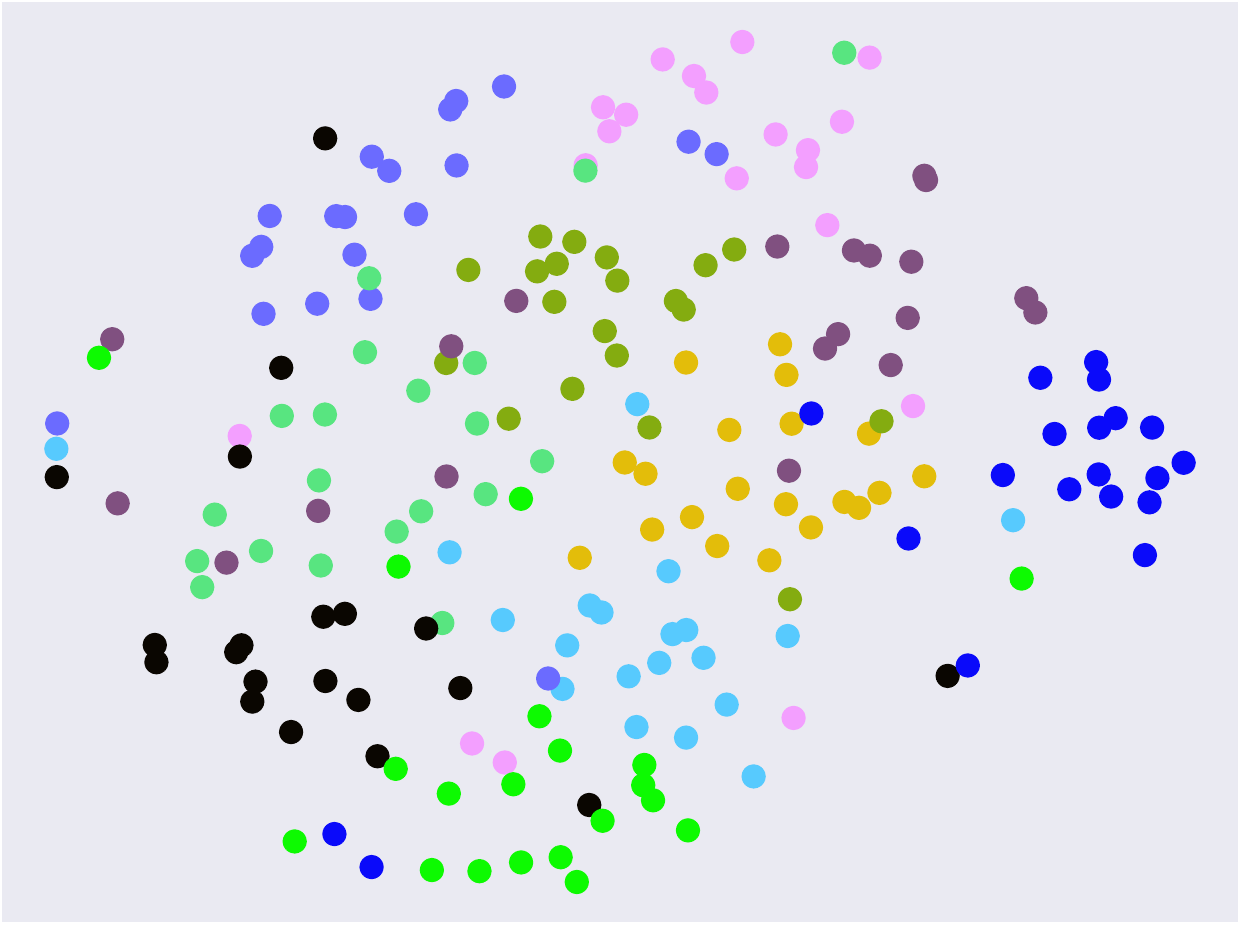} &     \includegraphics[height=0.085\textheight]{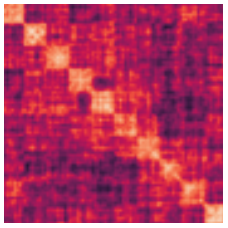} &
     \includegraphics[height=0.085\textheight]{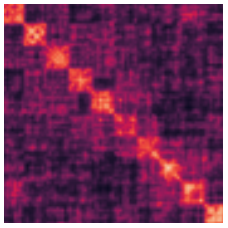} & 
     \includegraphics[height=0.085\textheight]{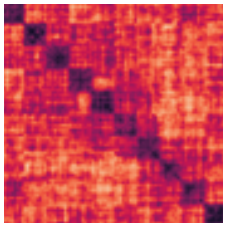} &
     \includegraphics[height=0.085\textheight]{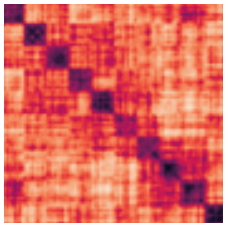} & 
     \includegraphics[height=0.085\textheight]{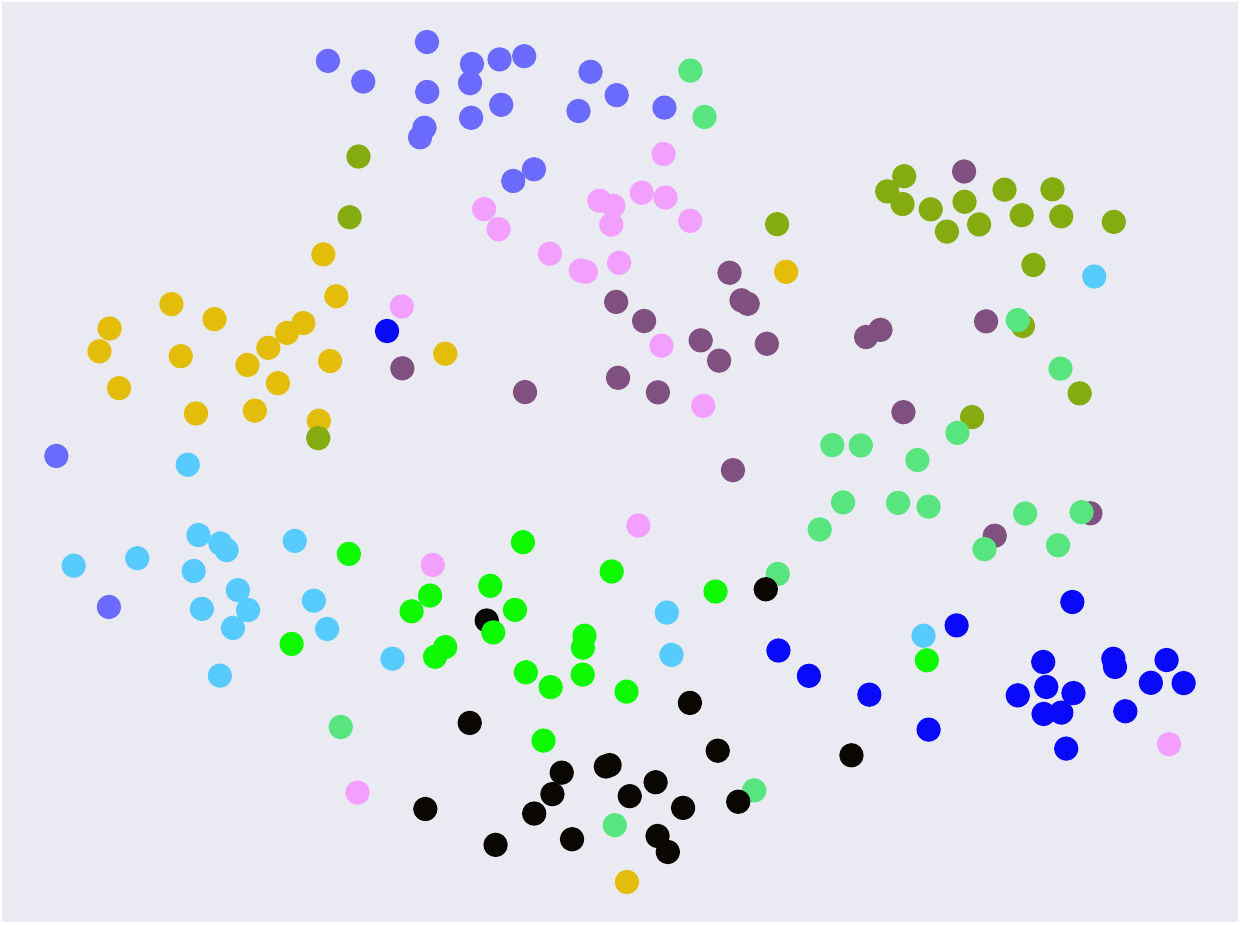} 
     \\
     (\textbf{a}) orig. & (\textbf{b}) $S$ & (\textbf{c}) $\mathcal{W}$ & 
     (\textbf{d}) $\mathcal{D}$ & (\textbf{e}) $\mathcal{D}_{\mathcal{W}}$ & (\textbf{f}) SOT
     \\
    \scriptsize{(orig. embedding)} & \scriptsize{(cos-similarity)} & \scriptsize{(SOT features)} & \scriptsize{$\;\;$(orig. dists)$\;\;$} &
    \scriptsize{$\;\;$(SOT dists)$\;\;$} & \scriptsize{(SOT embedding)}
    \\
\end{tabular}\\ \vspace{-7pt} ~\\
\caption{    {\fontsize{8.5}{8.5} \selectfont
\textbf{A close look at the SOT transform as it operates on a 10-way 20-shot supervised clustering task}: The input is a set of 200 33-dimensional unit-length feature vectors that are visualized on the plane in (\textbf{a}) using a t-SNE dimension reduction \cite{van2008visualizing}, where colors refer to the 10 classes. In (\textbf{b}) is the pairwise cosine similarity matrix $\mathcal{S}$, which is linearly related to the Euclidean pairwise distances $\mathcal{D}$ shown in (\textbf{d}). 
Next, in (\textbf{c}) we show the SOT matrix $\mathcal{W}$ whose rows (or columns, symmetrically) consist of our new embedding of the features. These 200-dimensional features are shown again on the plane in (\textbf{f}). Notice the visually apparent improvement in point gathering by class, from (\textbf{a}) to (\textbf{f}), which can be explained by comparing the matrices $\mathcal{D}$ and $\mathcal{D}_{\mathcal{W}}$, which are the self-pairwise distances of the original and SOT embedding respectively. Notice the greater contrast in $\mathcal{D}_{\mathcal{W}}$ between inter- and intra- cluster points. Note, that like in the visualizations of Fig.~\ref{fig.SOT}, we show the matrices with row/col order based on the true classes, purely for ease of visualization.}}
    \label{fig.close_look}\vspace{-10pt}
\end{figure}

The optimization problem in \cref{eq.opt_fractional_matching} is in fact a fractional matching instance between the set of $n$ original features and itself. It can be posed as a bipartite-graph min-cost max-flow instance. The graph has $n$ nodes on each side, representing the original features $\{v_i\}_{i=1}^n$ (the rows of $\mathcal{V}$). Across the two sides, the cost of the edge $(v_i,v_j)$ is the distance $d_{ij}$ and the edges of the type $(v_i,v_i)$ have a cost of infinity (or can simply be removed). Each `left' node is connected to a 'source' node by an edge of cost $0$ and similarly each 'right' node is connected to a `target' (sink) node by an edge of cost $0$. All edges in the graph have a capacity of $1$ and the goal is to find an optimal fractional self matching, by finding a min-cost max-flow from source to sink. Note that the maximum flow can easily be seen to be $n$, but a min-cost flow is sought among the max-flows. 

In this set-to-itself matching view, each vector $v_i$ is fractionally matched to the set of all other vectors $\mathcal{V}-\{v_i\}$  based on the pairwise distances, but importantly taking into account the fractional matches of the rest of the vectors in order to satisfy the double-stochasticity constraint\footnote{The construction constrains the maximum flow to exactly have a total outgoing flow of 1 from each `left' node and a total incoming flow of 1 from each `right' node.}.
Therefore, the $i$th transformed (re-embedded) feature $w_i$ ($i$th row of $\mathcal{W}$) is a \textit{distribution} (non-negative entries, summing to $1$), where $w_{ii}=0$ and  $w_{ij}$ is the relative belief that features $i$ and $j$ belong to the same `class'.

\begin{figure}[t]
\vspace{12pt}
\setlength{\tabcolsep}{6pt} 
\begin{tabular}{cc}
\hspace{-17pt}
\begin{minipage}{.42\textwidth}
\vspace{-10pt}
\includegraphics[width=1\textwidth]{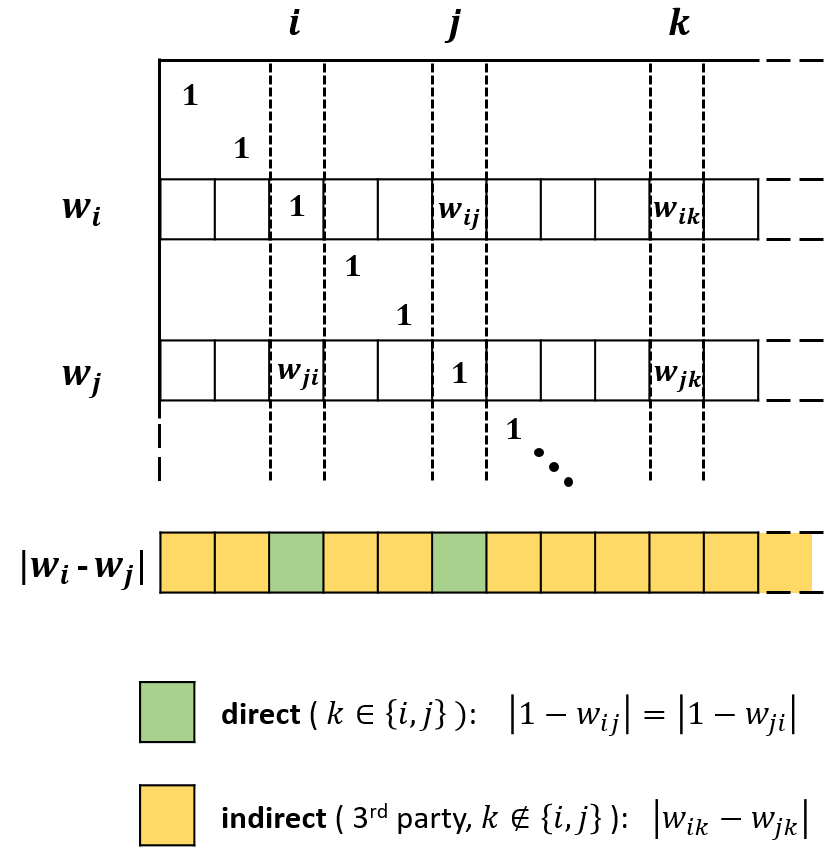}
\end{minipage}
     &  
\begin{minipage}{.56\textwidth}
\caption{
{\fontsize{8.5}{8.5} \selectfont
{\textbf{The (symmetric) embedding matrix $\mathcal{W}$ and the absolute difference between its $i$th and $j$th rows}: We examine the vector $|w_i-w_j|$: (i) Its $i$th and $j$th coordinates equal $|1-w_{ij}|=|1-w_{ji}|$, giving the \textit{direct} similarity between the original features, since this amount (in green) is greater when $w_{ij}$ and $w_{ji}$ (the mutual beliefs) are high (closer to 1). ; (ii) Its $k$th coordinate (for any $k\notin\{i,j\}$) gives $|w_{ik}-w_{jk}|$ which is an \textit{indirect} (third-party) comparison between the original features through the $k$th feature. Similarity (in yellow) is stronger when features $i$ and $j$ have a similar belief regarding feature $k$, \textit{i.e.} $w_{ik}$ and $w_{jk}$ are close.}}}
\label{fig.embedded_differences}
\end{minipage}
\\
\end{tabular}
        \\ \vspace{-28pt} \\
\end{figure}

Our final set of features $\mathcal{W}$ is obtained by replacing the diagonal entries from $0$s to $1$s, namely $\mathcal{W}=\mathcal{W}+I$, where $I$ is the $n\times n$ identity matrix.
Please refer to Fig.~\ref{fig.close_look} for a close look at the application of SOT to a toy clustering problem, where we demonstrate visually the improved embedding obtained through examining the pairwise distances before and after the transform.
We can now point out some important properties of this new embedding $\mathcal{W}$:

\vspace{3pt}\noindent \textbf{Direct and Indirect similarity encoding}: Each embedded feature encodes its distribution of similarities to all other features. An important property of our embedding is that the comparison of the embedded vectors $w_i$ and $w_j$ includes both \textit{direct} and \textit{indirect} information about the similarity between the features. Please refer to \cref{fig.embedded_differences} for a detailed explanation of this property. If we look at the different coordinates $k$ of the absolute difference vector $a=|w_i-w_j|$, SOT captures
(i) \textit{direct similarity}: For $k$ which is either $i$ or $j$, it holds that $a_k=1-w_{ij}=1-w_{ji}$ \footnote{Note: (i) $w_{ii}=w_{jj}=1$ ; (ii) $w_{ij}=w_{ji}$ from the symmetry of $\mathcal{W}$ ; (iii) all elements of $\mathcal{W}$ are $\leq 1$ and hence the $|\cdot|$ can be dropped ;}. This amount measures how high (\ie close to $1$) is the mutual belief of features $i$ and $j$ about one another.
(ii) \textit{indirect (3rd-party) similarity}: For $k\notin\{i,j\}$, we have            $a_k=|w_{ik}-w_{jk}|$, which is a comparison of the beliefs of features $i$ and $j$ regarding the (third-party) feature $k$.


\vspace{3pt}\noindent \textbf{Parameterless-ness}:
    Our proposed transform is parameterless, giving it the flexibility to be used in other pipelines, directly over different kinds of embeddings, without the harsh requirement of retraining the entire pipeline\footnote{Retraining is certainly possible, and beneficial in many situations, but not mandatory, as our experiments work quite well without it.}. 

\vspace{3pt}\noindent \textbf{Differentiability}:
    Due to the differentiability of Cuturi's \cite{cuturi2013sinkhorn} version of Sinkhorn, back-propagating through the SOT can be done naturally, hence it is possible to (re-)train the hosting network to adapt to the SOT, if desired.

\vspace{3pt}\noindent \textbf{Equivariance}: 
    The embedding works coherently with respect to any change of order of the input items (features). This can be shown by construction, since  min-cost max-flow solvers as well as the Sinkhorn OT solver are equivariant with respect to permutations of their inputs.

\vspace{3pt}\noindent \textbf{Explainability}:
    The non-parametric nature gives SOT an advantage over other set-to-set methods such as transformers in that its output is interpretable (e.g. by visually inspecting the transport-plan matrix $\mathcal{W}$), with a clear probabilistic characterization of the relations it had found.

\vspace{3pt}\noindent \textbf{Task-Aware Dimensionality}: SOT has the unique property that the dimension of the embedded feature depends on (equals) the number of features. On the one hand, this is a desired property, since it is only natural that the feature dimensionality (capacity) depends on the complexity of the task, which typically grows with the number of features (think of the inter-relations which are more complex to model). On the other hand, it might impose a problem in situations in which the downstream calculation that follows the feature embedding expects a fixed input size, for example a pre-trained non-convolutional layer.
Nevertheless, in many situations the downstream computation has the flexibility to work with varying input dimensions. Also, in most benchmarks the instance set sizes are fixed, allowing for a single setting of sizes to work throughout.

%% file: implementation_details.tex
\noindent\textbf{Datasets:} We consider three different applications to evaluate the performance of our method. 
For \textit{unsupervised clustering} we designed a specialized synthetic data set with the goal of enabling controlled experiments over a wide range of difficulties, which are determined by data dimensionality and in-cluster spread.

For \emph{few-shot classification} we use the standard benchmarks in the literature. The \emph{MiniImagenet} \cite{vinyals2016matching} dataset is a subset of \emph{Imagenet} \cite{Imagenet} that contains 100 classes and 600 images of size 84x84 per class. We follow the standard setup of using 64 classes for training and 16 and 20 novel classes for validation and testing. The \emph{CIFAR-FS} \cite{CIFAR} dataset includes 100 classes with 600 images of size 32 × 32 per-class. We used the same splits as in \emph{MiniImagenet} for this dataset. The \emph{CUB} \cite{CUB} dataset includes 200 classes of bird species and has 11,788 images of size 84 × 84 pixels in total. We followed the split suggested in \cite{CUBSPLIT} into 100 base classes, 50 validation classes and 50 novel classes.

For \textit{person re-identification} (ReID) we use two common large-scale datasets. The \emph{Market-1501} \cite{Market} and \emph{CUHK03} \cite{CUHK03} dataset consists of 1,501 and 1,467 identities and a total of 32,668 and 14,097 images taken from 6 cameras. 
We use the validation and test sets according to the splits in \cite{torchreid}.

\noindent\textbf{Pre-training:} We pre-trained ProtoNet~\cite{snell2017prototypical} with a 4-layer Convolution network adapting the procedures of \cite{snell2017prototypical} for training both with and without SOT, training on a 5-way (5/1)-shot 15-query task, using ADAM \cite{kingma2014adam} with learning rate 0.01 and step size of 20 over 100 episodes (tasks) per epoch.

\noindent\textbf{Fine-tuning:} We perform fine-tuning on two types of backbone residual networks - a resnet-12 as used in \cite{ye2020few} and a WRN-28-10 as used in \cite{mangla2020charting}.
For ProtoNet~\cite{snell2017prototypical} and ProtoNet-SOT, we fine-tune the base network with parameters taken from \cite{ye2020few}.
For PTMAP-SOT, we use meta-training with batches of a single 10-way 5-shot 15-query task per batch. We use ADAM with learning rate $5e-5$ that decreases with step size 10 for 25 epochs. We train the WRN-28-10 and the resnet-12 backbones for 800 and 100 episodes respectively per epoch.

\noindent\textbf{Hyper-parameters}: SOT has two hyper-parameters which were chosen through cross-validation and were kept fixed for each of the  applications over all datasets. (i) The number of Sinkhorn iterations for computing the optimal transport plan was fixed to 10. (ii) The entropy regularization parameter $\lambda$  (Eq. \eqref{eq:sinkhorn_objective}) was set to 0.1 for clustering and few-shot-learning experiments and to 1.0 for the ReID experiments. We further ablate these in the supplementaries.


%% file: results.tex
\newcommand{\mc}{\multicolumn{2}{|c|}}

\begin{figure}[t]
\hspace{-8pt}
\vspace{0pt}
\setlength{\tabcolsep}{5pt} 
\begin{tabular}{c c}
\begin{minipage}{3.6cm}
    \setlength{\tabcolsep}{-0.3em} 
\begin{tabular}{c c c}
     \includegraphics[width=1.4cm]{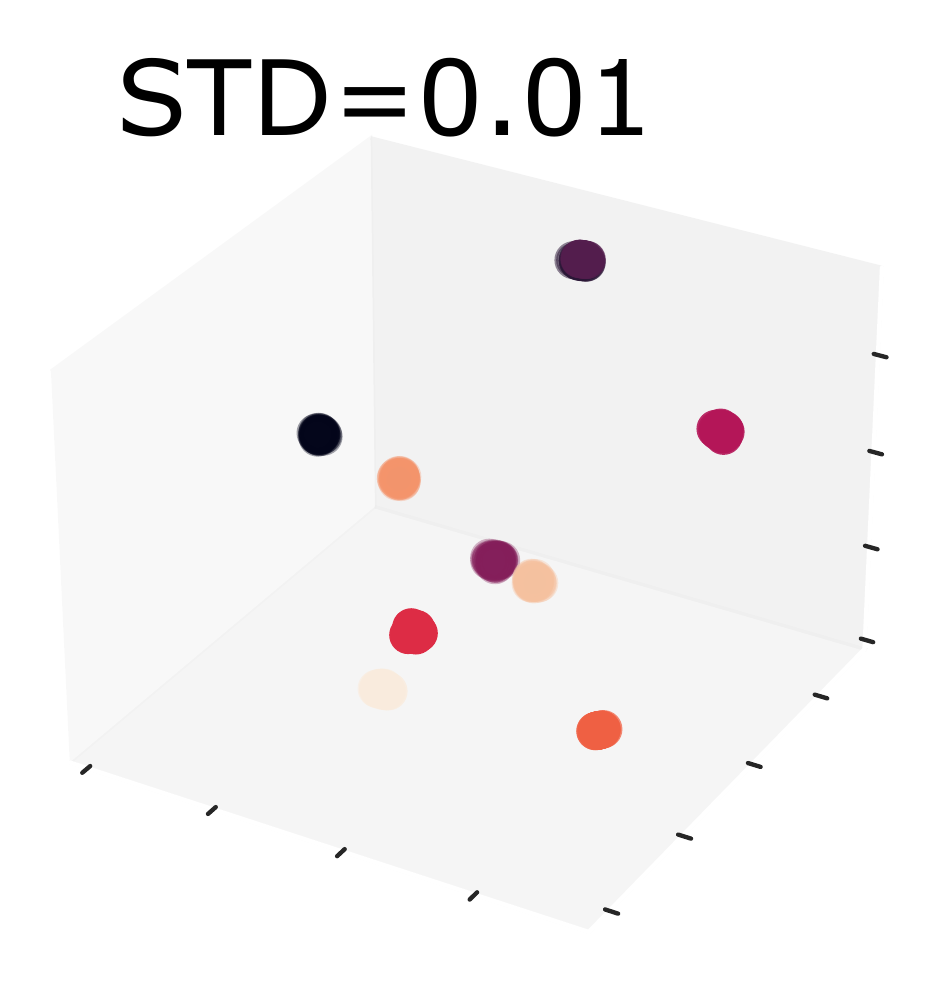} &
     \includegraphics[width=1.4cm]{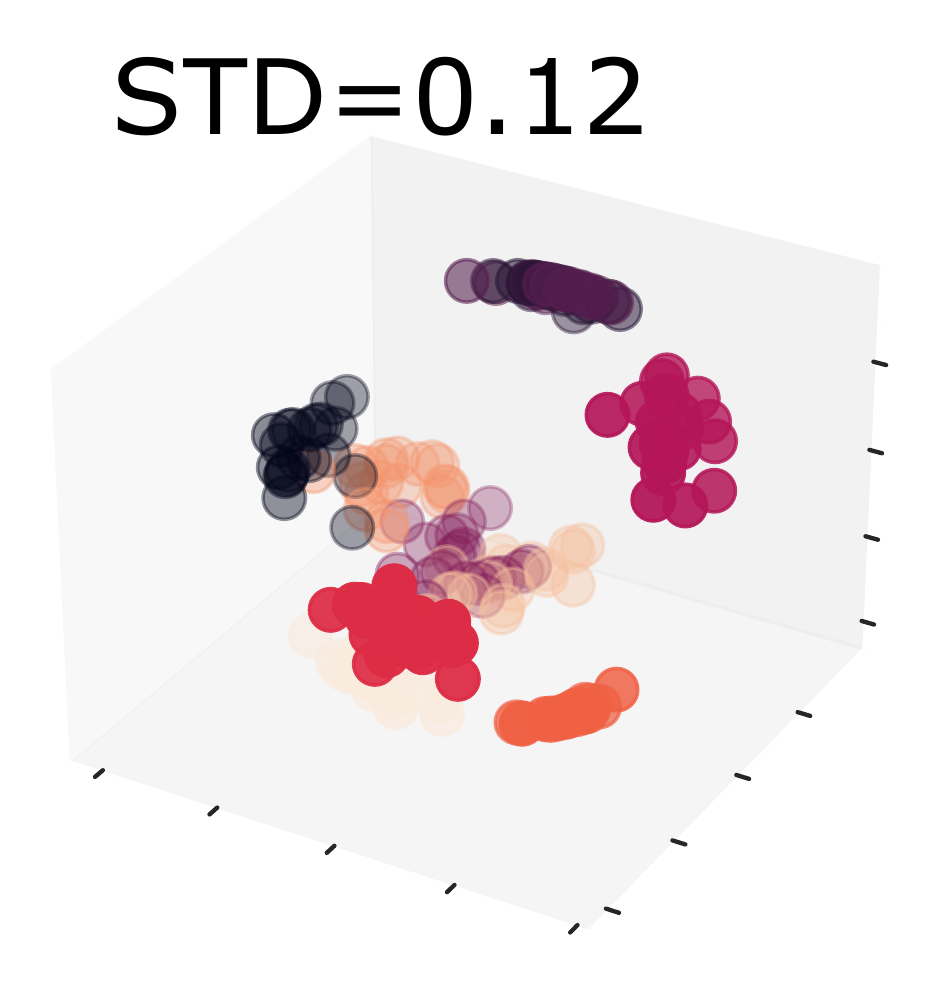} &
     \includegraphics[width=1.4cm]{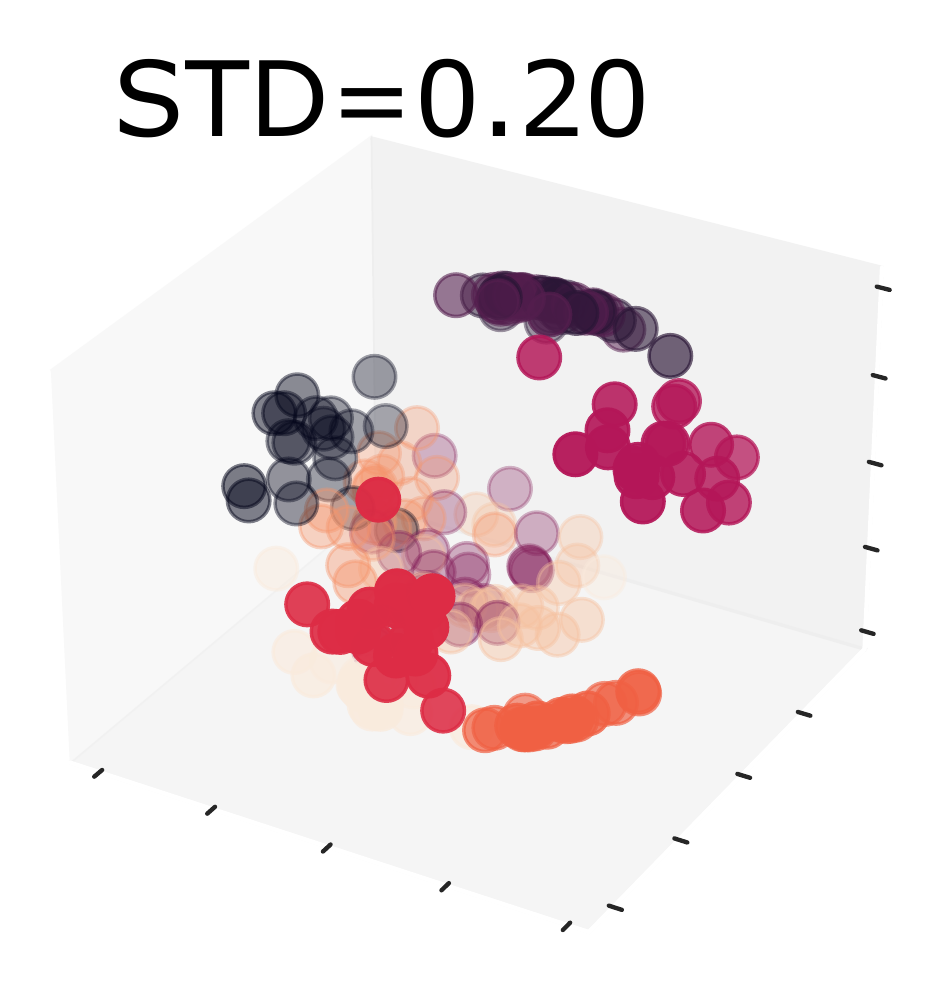} \\
     \includegraphics[width=1.4cm]{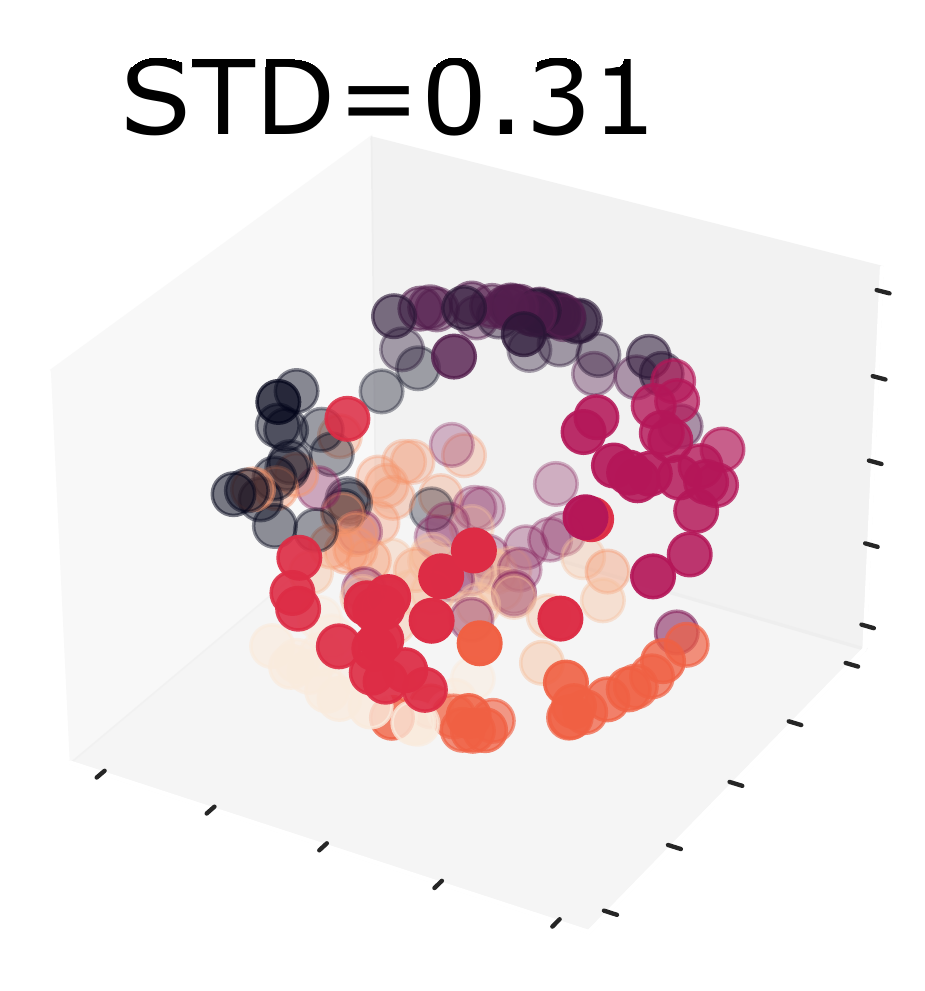} &     
     \includegraphics[width=1.4cm]{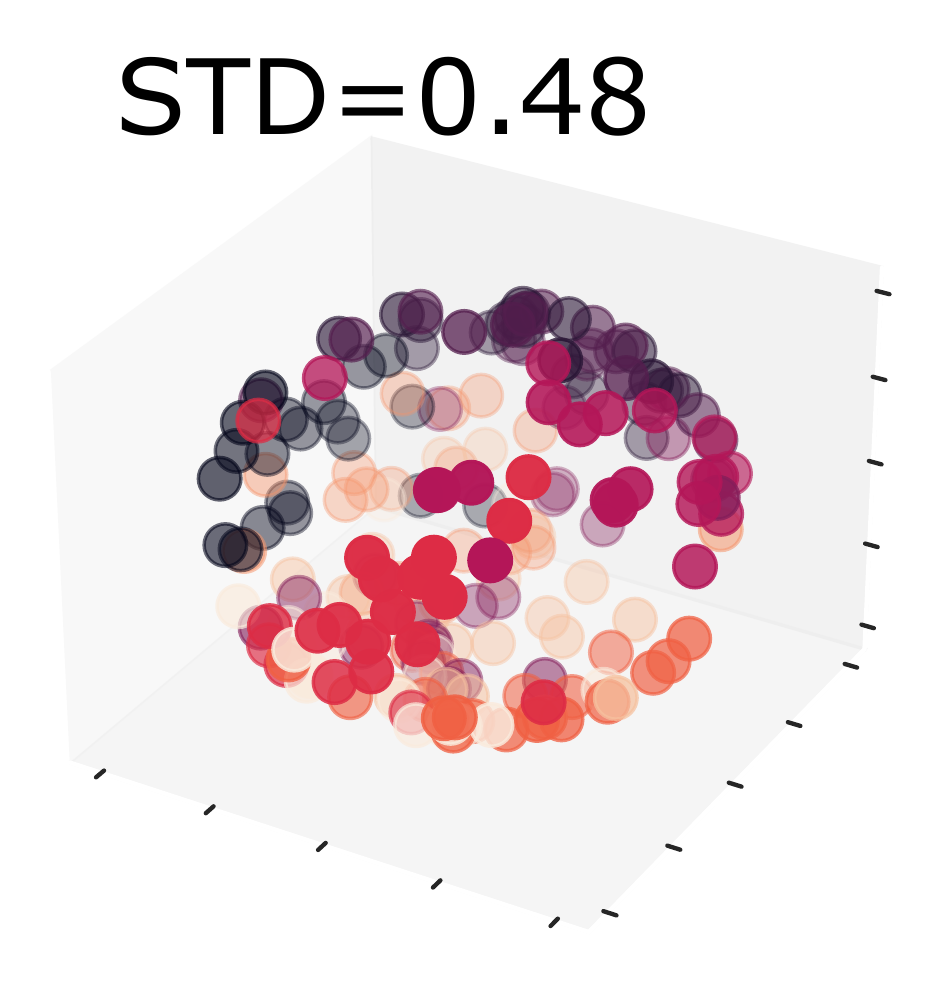} &
     \includegraphics[width=1.4cm]{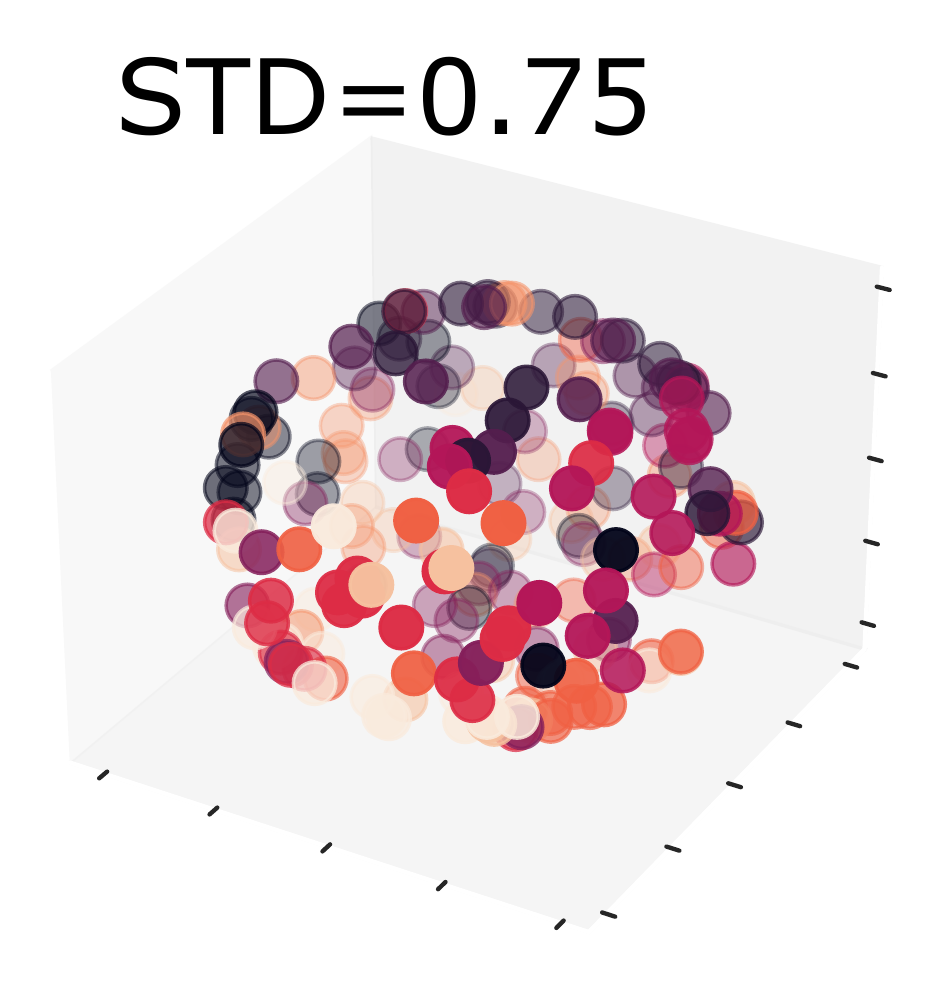} \\\vspace{-7pt}
\end{tabular}\\
{\fontsize{7}{7} \selectfont
(\textbf{i}) 10 Random cluster centers on the unit sphere, perturbed with increasing noise STD $\sigma$.}
\end{minipage}
&
\begin{minipage}{8.2cm}
~\\\vspace{-3pt}~\\
 \includegraphics[width=8.3cm]{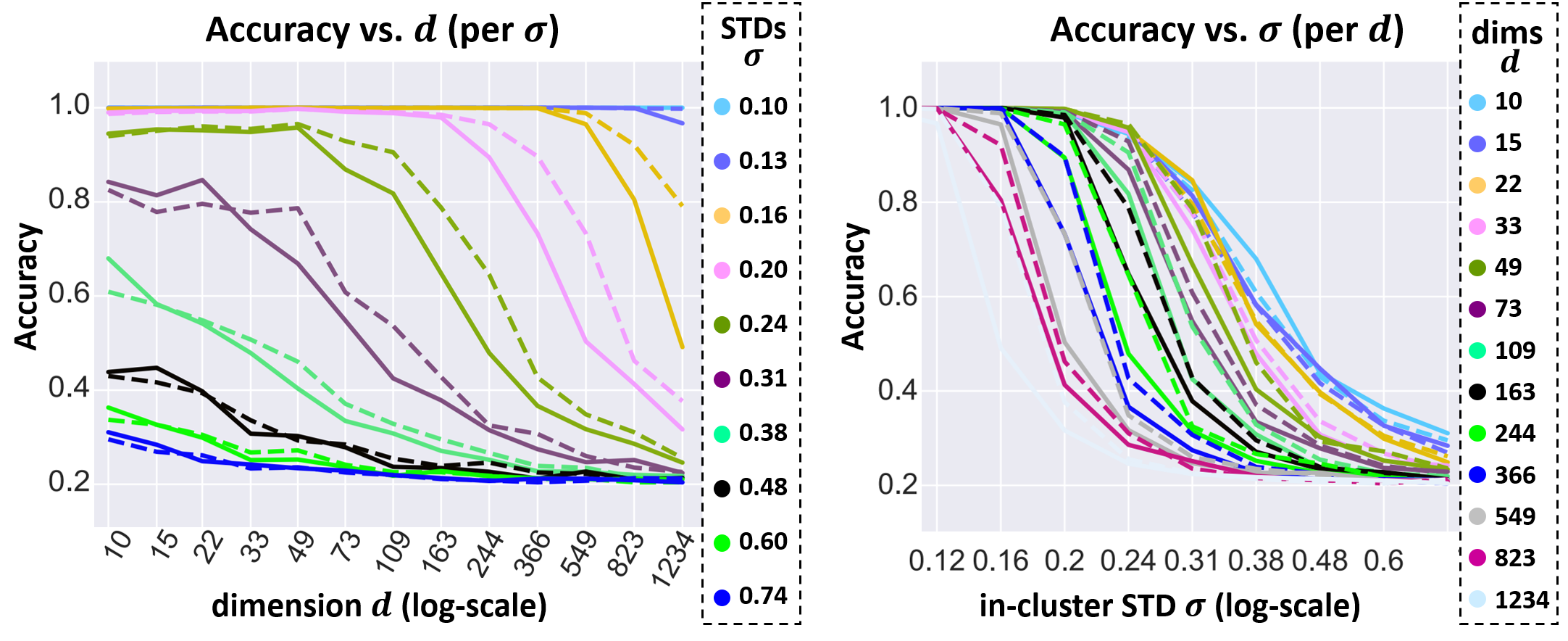} \\ \vspace{-15pt} ~\\
 \hspace{61pt}
{\fontsize{7}{7} \selectfont
(\textbf{ii}) Clustering accuracy across dimensions $d$ (left) and noise levels $\sigma$ (right). For each configuration, k-means accuracy is reported when applied with original (\textit{solid}) and SOT (\textit{dashed}) features.}
\end{minipage}
\end{tabular}
\\ \vspace{-16pt} \\
   \vspace{0pt}\caption{
   {\fontsize{8.5}{8.5} \selectfont
\textbf{Clustering on the \textit{d}-dimensional sphere}. \textbf{Left} (\textbf{i}): the data generation process (illustrated for the 3D case). \textbf{Right} (\textbf{ii}): detailed k-means accuracy results. The SOT (dashed) features give superior results throughout a majority of the space of settings.
       }}
    \label{fig.synthetic_exp}\vspace{-11pt}
\end{figure}

\subsection{Clustering on the Sphere}
We first demonstrate the effectiveness of SOT using a controlled synthetically generated clustering experiment, with $k=10$ cluster centers that are distributed uniformly at random on a $d$-dimensional unit-sphere, and 20  points per cluster (200 in total) that are perturbed around the cluster centers by Gaussian noise of increasing standard deviation, of up to 0.75, followed by a re-projection back to the sphere by dividing each vector by its $L_2$ magnitude. We also apply dimensionality reduction with PCA to $d=50$, for dimensions above 50.

We performed the experiment over a logarithmic 2D grid of combinations of data dimensionalities $d$ in the range $[10, 1234]$ and Gaussian in-cluster noise STD in the range $[0.1,0.75]$. Refer to Fig.~\ref{fig.synthetic_exp} (i) for a visualization of the data generation process. 

Each point is represented by its $d$-dimensional euclidean coordinates vector, where the baseline clustering is obtained by running k-means on these location features. In addition, we run k-means on the set of features that has undergone SOT. Hence, the benefits of the transform (embedding) are measured indirectly through the accuracy\footnote{Accuracy is measured by comparison with the optimal permutation of the predicted labels, found by the Hungarian Algorithm~\cite{kuhn1955hungarian}.} achieved by running k-means on the embedded vs. original vectors.
Evaluation results are reported in Fig.~\ref{fig.synthetic_exp} (ii) as averages over 10 runs, by plotting accuracy vs. dimensionality (for different noise STDs) and accuracy vs noise STDs (for different dimensionalities). The results show (i) general accuracy gains and robustness to wide ranges of data dimensionality (ii) the ability of SOT to find meaningful representations that enable clustering quality to degrade gracefully with the increase in cluster noise level. Note that the levels of noise are rather high, as they are relative to a unit radius sphere (a 3-dimensional example is shown at the top of the figure). We provide further details on this experiment in the supplementaries.

\newcommand{\bbb}[1]{({\footnotesize \color{blue} #1})}
\newcommand{\rrr}[1]{({\footnotesize \color{red} #1})}
\newcommand{\cmark}{\hspace{-10pt}\ding{51}\hspace{-6pt}}%
\newcommand{\xmark}{\hspace{-10pt}\ding{55}\hspace{-6pt}}%
\begin{table}[t]
    \centering
    \small
    \setlength{\tabcolsep}{0.75em} 
    \begin{tabular}{lcccc}
    \hline    
    \textit{method}    &  \hspace{-22pt}\textit{transductive}\hspace{-5pt}    &  \footnotesize{\textit{backbone}}  &  \footnotesize{5way-1shot}  &  \footnotesize{5way-5shot}   \\\hline 
    MAML(*)        \cite{finn2017model}& \xmark  & conv-4    &  46.47  &  62.71  \\ 
    RelationNet(*) \cite{sung2018learning}    & \xmark  & conv-4    &  49.31  &  66.60  \\ 
    ProtoNet(\#) \cite{snell2017prototypical} & \xmark  & conv-4    &  49.10  &  66.79  \\ 
    FEAT(\$) \cite{ye2020few}   & \xmark  & conv-4 &  \textbf{55.15}  &  \textbf{71.61}  \\ 
\hdashline
    ProtoNet-SOT$\blue{_\textit{p}}$  & \cmark  & conv-4    &  \underline{{54.01}} \bbb{ +10.2\%}  &  69.39 \bbb{+3.9\%}  \\   
    ProtoNet-SOT$\red{_\textit{t}}$  & \cmark  & conv-4    &  53.70 \rrr{+9.3\%}  &  \underline{{70.40}} \rrr{+5.4\%}  \\    
\hline
    ProtoNet(\#) \cite{snell2017prototypical} & \xmark  & resnet-12 &  62.39  &  80.33  \\ 
    DeepEMD(\$) \cite{DeepEMD}    & \xmark  & resnet-12 &  65.91  &  82.41  \\
    FEAT(\$)    \cite{ye2020few}  & \xmark  & resnet-12 &  66.78  &  82.05  \\
    RENet(\$)   \cite{ReNet}      & \xmark  & resnet-12 &  67.60  &  82.58  \\
    PTMAP(\#) \cite{hu2020leveraging}  & \cmark  & resnet-12 &  76.90  &  85.20  \\ 
\hdashline
    ProtoNet-SOT$\blue{_\textit{p}}$  & \cmark  & resnet-12 &  67.34  \bbb{+7.9\%}  & 81.84 \bbb{+1.6\%} \\   
    ProtoNet-SOT$\red{_\textit{t}}$  & \cmark  & resnet-12 &  67.90 \rrr{+8.8\%}  &  83.09 \rrr{+3.2\%}  \\
    PTMAP-SOT$\blue{_\textit{p}}$     & \cmark  & resnet-12 &  \textbf{78.35} \bbb{+1.9\%}  &  \textbf{86.01} \bbb{+1.0\%}  \\   
    PTMAP-SOT$\red{_\textit{t}}$     & \cmark  & resnet-12 &  \underline{{77.30}} \rrr{+0.5\%}  &  \underline{{85.49}} \rrr{+0.3\%}  \\    
    \hline

    ProtoNet(\&) \cite{snell2017prototypical} & \xmark  & wrn-28-10 &  62.60  &  79.97  \\ 
    PTMAP(\$) \cite{hu2020leveraging}  & \cmark  & wrn-28-10 &  82.92  &  88.80  \\ 
    Sill-Net(\$) \cite{zhang2021sill}  & \cmark  & wrn-28-10 &  82.99  &  89.14  \\ 
    PTMAP-SF(\$) \cite{chen2021few}    & \cmark  & wrn-28-10 &  \underline{{84.81}}  &  \underline{{90.62}}  \\ 
\hdashline
    PTMAP-COSINE     & \cmark  & wrn-28-10 &  74.60 \bbb{-10.0\%} &  84.68 \bbb{-4.6\%} \\   
    PTMAP-SOFTMAX    & \cmark  & wrn-28-10 &  80.08 \bbb{-3.4\%} &  83.83 \bbb{-5.6\%} \\
\hdashline


    PTMAP-SOT$\blue{_\textit{p}}$     & \cmark  & wrn-28-10 &  83.19 \bbb{+0.3\%} &  89.56 \bbb{+0.9\%} \\   
    PTMAP-SOT$\red{_\textit{t}}$     & \cmark  & wrn-28-10 &  84.18 \rrr{+1.5\%} &  90.51 \rrr{+1.9\%} \\
    Sill-Net-SOT$\blue{_\textit{p}}$  & \cmark  & wrn-28-10 &  83.35 \bbb{+0.4\%} &  89.65 \bbb{+0.6\%} \\
    PTMAP-SF-SOT$\blue{_\textit{p}}$  & \cmark  & wrn-28-10 &  \textbf{85.59} \bbb{+0.9\%}  &  \textbf{91.34}  \bbb{+0.8\%} \\
\hline           
    \end{tabular} \vspace{3pt}
    \caption{
    {\fontsize{8.5}{8.5} \selectfont
    \textbf{Few-Shot Classification (FSC)} accuracy on \textbf{MiniImagenet} \cite{vinyals2016matching}. The improvements introduced by the variants of SOT (percentages in brackets) are in comparison with each respective baseline hosting method. \textbf{Bold} and \underline{underline} notations highlight best and second best results per backbone.
         (*) = from \cite{chen2018closer} ;
        (\&) = from \cite{ziko2020laplacian} ; 
        (\$) = from the method's paper itself ; 
        (\#) = our implementation ;
        }
        }
        ~\\ \vspace{-35pt} ~\\
    \label{tab:results_fsc_MiniImagenet}
\end{table} 

\subsection{Few-Shot Classification (FSC)}
Our main experiment is a comprehensive evaluation on the standard few-shot classification benchmarks \emph{MiniImagenet} \cite{vinyals2016matching}, \emph{CIFAR-FS} \cite{CIFAR}, and \emph{CUB} \cite{CUB}, with detailed results in Tables \ref{tab:results_fsc_MiniImagenet} and \ref{tab:results_fsc_CIFAR_CUB}. 
For \emph{MiniImagenet} (Table \ref{tab:results_fsc_MiniImagenet}) we report on both versions ``SOT$\blue{_\textit{p}}$" and ``SOT$\red{_\textit{t}}$" over a range of backbone architectures, while for the smaller datasets \emph{CIFAR-FS}  and \emph{CUB} (Table \ref{tab:results_fsc_CIFAR_CUB}) we focus on the `drop-in' version ``SOT$\blue{_\textit{p}}$" and only the strongest wrn-28-10 architecture.

One goal here is to show that we can achieve new state-of-the-art FSC results, when we build on current state-of-the-art. But more importantly, we demonstrate the flexibility and simplicity of applying SOT in this setup, with improvements in the entire range of testing, including: (i) when building on different `hosting' methods; (ii) when working above different feature embeddings of different complexity backbones; and (iii) whether retraining the hosting network or just dropping-in SOT and performing standard inference.

To evaluate the performance of the proposed SOT, we applied it to previous FSC methods including the very recent state-of-the-art (PT-MAP \cite{hu2020leveraging}, Sill-NET \cite{zhang2021sill} and PT-MAP-SF \cite{chen2021few}) as well as a to more conventional methods like the popular ProtoNet \cite{snell2017prototypical}. The detailed results are presented in Tables \ref{tab:results_fsc_MiniImagenet} and \ref{tab:results_fsc_CIFAR_CUB}) for the different datasets. Note that SOT is by nature a transductive method\footnote{SOT is transductive in the sense that it needs to jointly process the data, but importantly, unlike other methods it does not gain its benefit in being so from making limiting assumptions about the structure of the instance, like knowing the number of classes, or the number of items per class.}, hence we marked its results as so, regardless of whether the hosting network is transductive or not. In the following, we discuss the two modes in which our transform can be used in existing FSC methods.



\vspace{3pt}\noindent\textbf{SOT insertion \textit{without} network retraining}
(notated by SOT$\blue{_\textit{p}}$ in Tables \ref{tab:results_fsc_MiniImagenet} and \ref{tab:results_fsc_CIFAR_CUB}).
Recall that the proposed transform is \textit{non-parametric}. As such, we can simply apply it to a trained network at inference, without the need to re-train. This basic `drop-in' use of SOT consistently, and in many cases also significantly, improved the performance of the tested methods, including state-of-the-art, across all benchmarks and backbones. 
SOT$\blue{_\textit{p}}$ gave improvements of around $3.5\%$ and $1.5\%$ on 1 and 5 shot \emph{MiniImagenet} tasks. 
This improvement without re-training the embedding backbone network shows SOT's effectiveness in capturing meaningful relationships between features in a very general sense. 

\newcommand{\bbbbb}[1]{{\footnotesize ({\color{blue} #1})}}
\newcommand{\rrrrr}[1]{({\footnotesize \color{red} #1})}
\newcommand{\foot}[1]{{\footnotesize #1}}

\begin{table}[t]
    \centering 
    \small
    \setlength{\tabcolsep}{0.3em} 
    \begin{tabular}{l:cc:cc}
    \hline  
    \textit{FSC benchmark}
    & \multicolumn{2}{c:}{\textit{CIFAR-FS}~\cite{CIFAR}} 
    & \multicolumn{2}{c}{\textit{CUB}~\cite{CUB}} \\
    \hline    
    \textit{method}     &  \footnotesize{5way-1shot}  &  \footnotesize{5way-5shot}  &  \footnotesize{5way-1shot}  &  \footnotesize{5way-5shot}   \\\hline 
    \footnotesize{PTMAP(\$) \cite{hu2020leveraging}}   &  87.69  &  90.68  &  91.55  &  93.99\\ 
    \footnotesize{Sill-Net(\$) \cite{zhang2021sill}}   &  87.73  &  91.09  &  94.73  &  96.28 \\ 
    \footnotesize{PTMAP-SF(\$) \cite{chen2021few}}     &  \underline{{89.39}}  &  \underline{{92.08}}  &  \underline{{95.45}}  &  \underline{{96.70}} \\ 
\hdashline
    PTMAP-SOT$\blue{_\textit{p}}$      &  87.37  \bbb{-0.4\%} &  91.12  \bbb{+0.5\%} & 91.90 \bbb{+0.4\%} &  94.63 \bbb{+0.7\%}  \\   
    Sill-Net-SOT$\blue{_\textit{p}}$   &  87.30  \bbb{-0.5\%} &  91.40 \bbb{+0.3\%} & 94.86 \bbb{+0.1\%} &  96.61 \bbb{+0.3\%} \\
    PTMAP-SF-SOT$\blue{_\textit{p}}$   &  \textbf{89.94} \bbb{+0.6\%}  &  \textbf{92.83} \bbb{+0.8\%}  &  \textbf{95.80} \bbb{+0.4\%} &  \textbf{97.12} \bbb{+0.4\%} \\
\hline           
    \end{tabular} \vspace{2pt}
    \caption{
    {\fontsize{8.5}{8.5} \selectfont
    \textbf{Few-Shot Classification (FSC)} accuracy on \textit{CIFAR-FS}~\cite{CIFAR} and \textit{CUB}~\cite{CUB}.
        }}         ~\\ \vspace{-35pt} ~\\

    \label{tab:results_fsc_CIFAR_CUB}
\end{table}

\vspace{3pt}\noindent\textbf{SOT insertion \textit{with} network retraining}
(notated by SOT$\red{_\textit{t}}$ in Table \ref{tab:results_fsc_MiniImagenet}).
Due to its \textit{differentiability} property, the proposed method can be applied while training and hence we expect an adaptation of the hosting network's parameters to the presence of the transform with a potential for improvement. 
To evaluate this mode, we focused on the \emph{MiniImagenet} benchmark \cite{vinyals2016matching}, specifically on the same configurations that we used without  re-training, to enable a direct comparison. The results in Table \ref{tab:results_fsc_MiniImagenet} show additional improvements in almost every method. SOT$\red{_\textit{t}}$ gave improvements of around $5\%$ and $3\%$ on 1 and 5 shot \emph{MiniImagenet} tasks, further improving on the pre-trained counterpart. This result indicates the effectiveness of training with SOT in an end-to-end fashion. 

\vspace{3pt}\noindent\textbf{Ablations} Within the context of few-shot learning on \emph{MiniImagenet}, we performed several ablation studies. In Table~\ref{tab:results_fsc_MiniImagenet}, the networks `PTMAP-COSINE' and `PTMAP-SOFTMAX' stand for the obvious baseline attempts (found to be unsuccessful) that work in the line of our approach, without the specialized OT-based transform. In the former, we take the output features to be the rows of the (un-normalized) matrix  $\mathcal{S}$ (rather than those of  $\mathcal{W}$) and in the latter we also normalize its rows using soft-max.
In the supplementaries we ablate on SOT's two parameters - the number of Sinkhorn iterations and the entropy term $\lambda$.

\subsection{Person re-Identification (Re-ID)}
In this section, we explore the possibility of using SOT on large-scale datasets by considering the Person re-Identification task. 
Given a set of \textit{query} images and a large set of \textit{gallery} images, the task is to rank the similarities of each single query against the gallery. This is done by computing specialized image features among which similarities are based on Euclidean distances. SOT is applied to such pre-computed image features, refining them with the strong relative information that it is able to capture by applying it on the union of all query and gallery features. We adapted a pre-trained standard resnet-50 architecture \cite{torchreid} and the popular TopDBNet \cite{Top-DB-Net}, which we tested on the large-scale ReID benchmarks \emph{CUHK03} \cite{CUHK03} (on the 'detected' version and similar results on the `labeled' version in the supplementaries) and \textit{Market-1501} \cite{Market}, with and without the re-ranking \cite{Re-ranking} procedure. For evaluation, we followed their conventions and compare results using the mAP (mean Average Precision) and Rank-1 metrics.

The results in Table \ref{tab:results_on_DukeMTMC_and_market} show a consistent benefit in using SOT within the different networks. For \emph{CUHK03}, the results improved by a large margin of $+6.8\%$ in $m$AP for the best configuration. These results demonstrate that the proposed SOT scales well to large-scale problems (with number of features in the  thousands) and is attractive for a variety of applications. 
ReID is not the main focus of this work, hence, we did not re-train the hosting networks with SOT included. Further research is required to measure the possible effects of doing so.

\begin{table}[t]
    \centering 
    \small
    \setlength{\tabcolsep}{0.25em} 
    \begin{tabular}{l | cc | cc}
    \hline
    \textit{ReID benchmark} & \multicolumn{2}{c}{\emph{CUHK03-detected} \cite{CUHK03}} & \multicolumn{2}{|c}{\emph{Market-1501} \cite{Market}}  \\
    \hline    
    \textit{network}    &  \foot{mAP}  &  \foot{Rank-1} &  \foot{mAP}  &  \foot{Rank-1}\\\hline
    TopDBNet \cite{Top-DB-Net}  & \foot{72.9}   &  \foot{75.7}    & \foot{85.7}   &  \foot{94.3}        \\
    TopDBNet-rerank \cite{Top-DB-Net} & \foot{\underline{87.1}}   &     \foot{\underline{87.1}}      &       \foot{\textbf{94.0}}   &    \foot{\textbf{95.3}}    \\

\hdashline

    TopDBNet-SOT$\blue{_\textit{p}}$  &       \foot{{{77.9}}} \bbbbb{+6.9\%}  &  \foot{80.4} \bbbbb{+6.2\%}    &     \foot{\underline{{88.1}}} \bbbbb{+2.8\%}     &         \foot{{94.4}} \bbbbb{+0.1\%}       \\
    TopDBNet-rerank-SOT$\blue{_\textit{p}}$     &   \foot{\textbf{87.9}} \bbbbb{+0.9\%}    &         \foot{\textbf{{88.0}}} \bbbbb{+1.0\%}  &    \foot{\textbf{94.0}} \bbbbb{0.0\%} &     \foot{\underline{{95.0}}} \bbbbb{-0.3\%} \\
\hline           
    \end{tabular} \vspace{2pt}
    \caption{
        {\fontsize{8.5}{8.5} \selectfont
        \textbf{Re-ID} results on \emph{CUHK03} \cite{CUHK03} and  \emph{Market-1501} \cite{Market}} } 
                ~\\ \vspace{-35pt} ~\\
    \label{tab:results_on_DukeMTMC_and_market}
\end{table}

%% file: conclusions.tex
\vspace{-4pt}
In this paper, we explored the idea of utilizing global information of features, for instance-specific problems such as clustering, few-shot learning, and person re-identification. We proposed a novel module: the Self-Optimal-Transport (SOT) - a features transform that is non-parametric, differentiable and which can capture high-level relationships between data points in problems of this nature. 
The proposed method outperforms state-of-the-art networks on popular few-shot classification benchmarks and shows consistent improvements on tested ReID benchmarks. Based on these promising results, we believe that exploring its full potential can lead to improvements in a variety of fields and open new possibilities.

In future work, we plan to address some current limitations. (i) Regarding the output dimensionality of the embedding, which is dictated by the input set size. We will aim at being able to obtain an arbitrary dimension, for increased usage flexibility; 
(ii) We plan to investigate the usage of SOT in unsupervised settings, which would be possible by utilizing its informative representation for self-supervision; (iii) It would likely be beneficial to have a variant of SOT in which the transform is enriched with learnable parameters, similar to transformers, to extend its modeling capacity even further; (iv) SOT is purely transductive. We plan to explore  non-transductive variants, possibly by comparing each sample separately to the support or gallery sets.

%% file: SOT_appendix.tex

\section{ablation studies}

\subsection{Sinkhorn iterations}
In Table \ref{tab:results_sinkhorn_iters} we ablate the number of normalization iterations in the Sinkhorn-Knopp (SK) \cite{cuturi2013sinkhorn} algorithm at test-time. We measured accuracy on the validation set of \emph{MiniImagenet} \cite{vinyals2016matching}, using ProtoNet-SOT$\blue{_\textit{p}}$ (which is the non-fine-tuned drop-in version of SOT within ProtoNet \cite{snell2017prototypical}). 
As was reported in prior works following \cite{cuturi2013sinkhorn}, we empirically observe that a very small number of iterations (around 5) provide rapid convergence. We observed similar behavior for other hosting methods, and therefore chose to use a fixed number of 10 iterations throughout the experiments.

\begin{table}[h!]
    \centering\vspace{-6pt}
    \small
    \setlength{\tabcolsep}{0.6em} 
    \begin{tabular}{lccc}
    \hline    
    \textit{method}    &  \footnotesize{iterations}  &  \footnotesize{5way-1shot}  &  \footnotesize{5way-5shot}   \\\hline 
    ProtoNet-SOT$\blue{_\textit{p}}$      &              1              &            70.71            &       83.79 \\ 
    ProtoNet-SOT$\blue{_\textit{p}}$      &              2              &            71.10            &       84.01 \\ 
    ProtoNet-SOT$\blue{_\textit{p}}$      &              4              &            71.18            &       84.08 \\ 
    ProtoNet-SOT$\blue{_\textit{p}}$      &              8              &            71.20            &       84.10 \\ 
    ProtoNet-SOT$\blue{_\textit{p}}$      &              16             &            71.20            &       84.10 \\    
    \end{tabular} \vspace{5pt}
    \caption{
      {\fontsize{9}{9} \selectfont
\textbf{Sinkhorn iterations ablation study:} See text for details.}} \vspace{-25pt}
    \label{tab:results_sinkhorn_iters}
\end{table}

\subsection{OT entropy regularization parameter $\lambda$} \label{sec.lambda}
\begin{figure}[h!]
\vspace{-12pt} \hspace{18pt}
\centering    \setlength{\tabcolsep}{-0.43em} 
\hspace{-29pt}\\\vspace{3pt}
 \includegraphics[height=4.5cm]{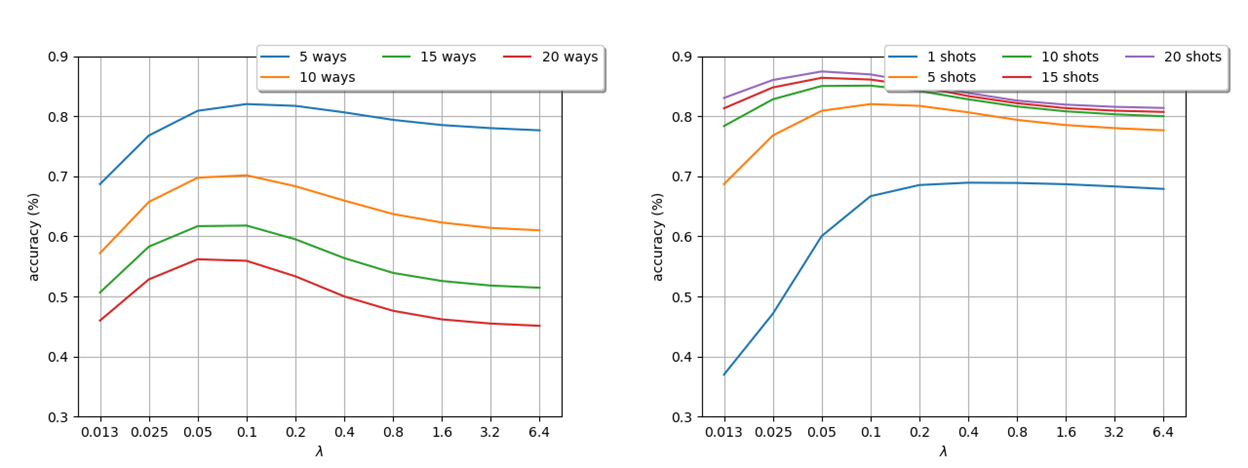} \\\vspace{-11pt}
  \vspace{0pt}\caption{
  {\fontsize{9}{9} \selectfont
  \textbf{Ablation study on $\lambda$ in \textit{Few-Shot-Classification} (FSC)}: Considering different `ways' (left), and different `shots' (right). See text for details.
  }}
    \label{fig.lambda_exp}\vspace{-3pt}
\end{figure}

\begin{figure}[h!]
\vspace{-0pt}
\setlength{\tabcolsep}{3pt} 
\begin{tabular}{cc}
\hspace{8pt}
\begin{minipage}{.4\textwidth}
\caption{
{\fontsize{9}{9} \selectfont
   \textbf{Ablation study on $\lambda$ in \textit{Person-Re-Identification} (Re-ID)}: Using the validation set of the Market-1501 dataset and considering both mAP and Rank-1 measures. See text for details.
   }}
\label{fig.lambda_exp_reid}
\end{minipage}
&
\begin{minipage}{.5\textwidth}
\vspace{-10pt}
\includegraphics[height=4.5cm]{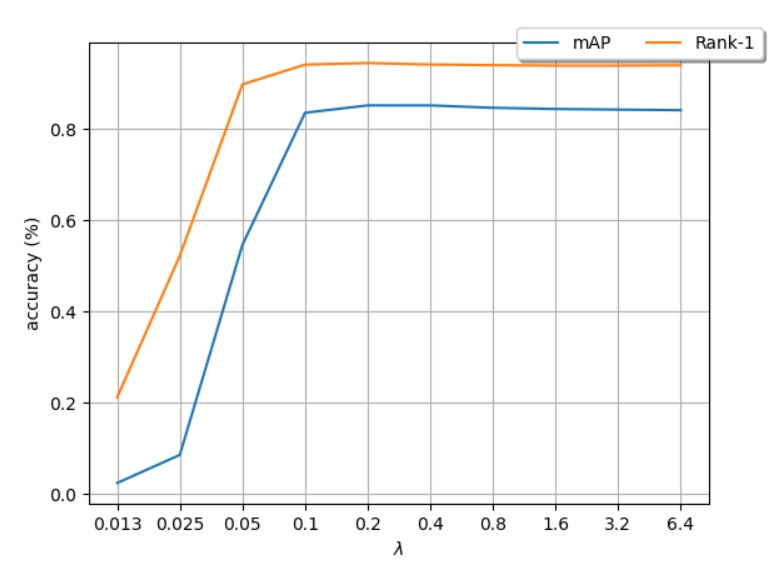}
\end{minipage}
\\
\end{tabular}
        \\ \vspace{-18pt} \\
\end{figure}

We measured the impact of using different values of the optimal-transport entropy regularization parameter $\lambda$ (the main parameter of the Sinkhorn algorithm) on a variety of configurations (ways and shots) in Few-Shot-Classification (FSC) on \emph{MiniImagenet} \cite{vinyals2016matching} in Fig. \ref{fig.lambda_exp} as well as on the Person-Re-Identification (RE-ID) experiment on Market-1501 \cite{Market} in Fig. \ref{fig.lambda_exp_reid}. In both cases, the ablation was executed on the validation set.

For FSC, in Fig. \ref{fig.lambda_exp}, the \textbf{left} plot shows that the effect of the choice of $\lambda$ is similar across tasks with a varying number of ways. The \textbf{right} plot shows the behavior as a function of $\lambda$ across multiple shot-values, where the optimal value of $\lambda$ can be seen to have a certain dependence on the number of shots. Recall that we chose to use a fixed value of $\lambda=0.1$, which gives an overall good accuracy trade-off. Note that a further improvement could be achieved by picking the best values for the particular cases. Notice also the log-scale of the x-axes to see that performance is rather stable around the chosen value.

For Re-ID, in Fig. \ref{fig.lambda_exp_reid}, we experiment with a range of $\lambda$ values on the validation set of the Market-1501 dataset. The results (shown both for mAP and rank-1 measures) reveal a strong resemblance to those of the FSC experiment in Fig. \ref{fig.lambda_exp}, however, the optimal choices for $\lambda$ are slightly higher, which is consistent with the dependence on the shots number, since the re-ID tasks are typically large ones. In this re-ID ablation, we found that a value of $\lambda=0.25$ gives good results across both datasets. We ask to note that in the paper we mistakenly reported that we used $\lambda=1.0$, while in practice all our results were obtained using $\lambda=0.25$.

%
%
%
%
%
%


\section{Unsupervised Clustering - further details}
In this section we provide further details (due to lack of space in main paper) on the experiment on unsupervised clustering on the unit sphere (Exp. 5.1).

\subsection{Separation between inter- and intra-class features}

\begin{figure}[t]
\centering    \setlength{\tabcolsep}{0.4em} 
\begin{tabular}{c c}
     \hspace{-5pt}
     \includegraphics[width=0.48\columnwidth]{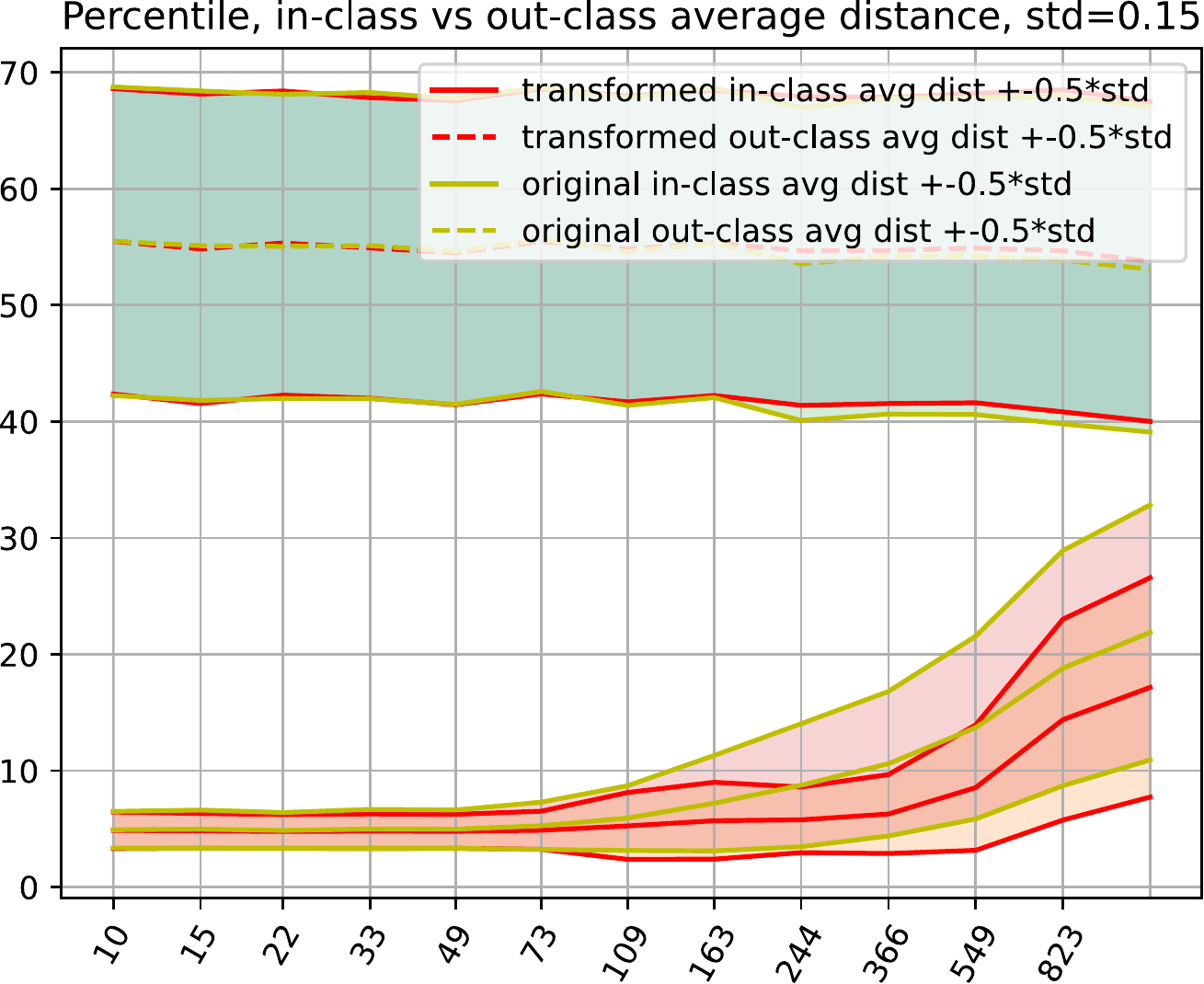} &
     \includegraphics[width=0.48\columnwidth]{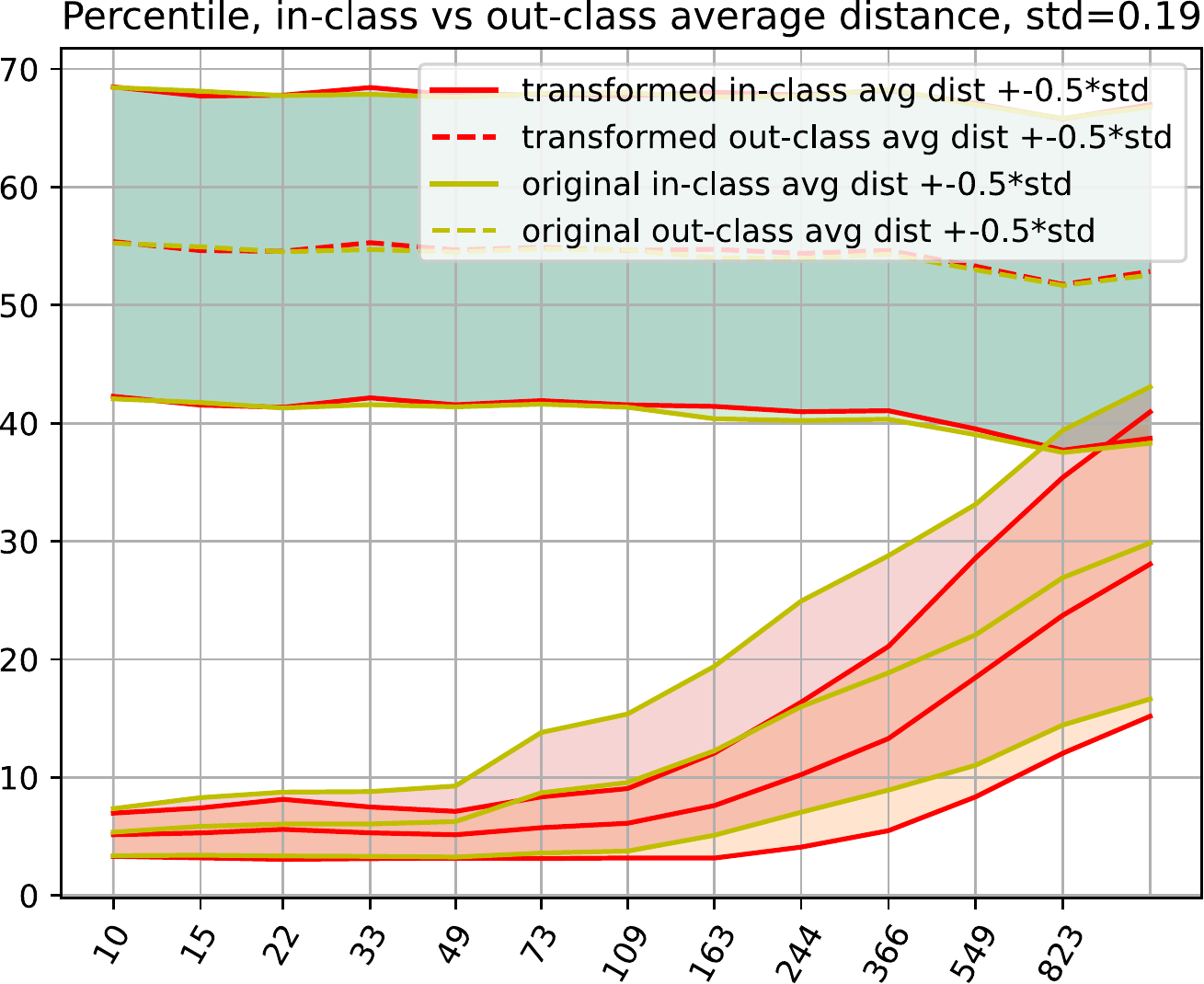} 
     \vspace{2pt}\\
     \hspace{-5pt}
     \includegraphics[width=0.48\columnwidth]{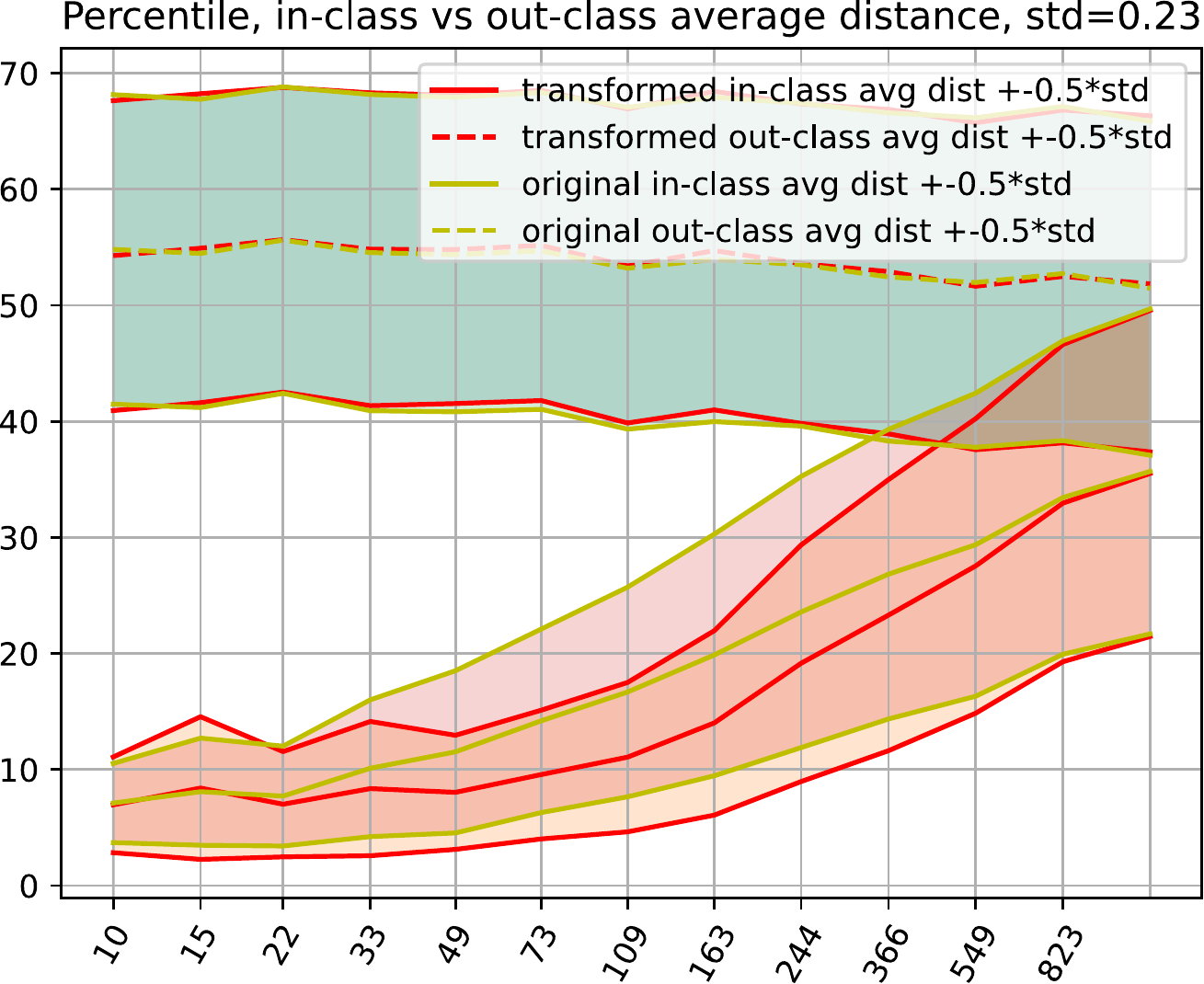} &
     \includegraphics[width=0.48\columnwidth]{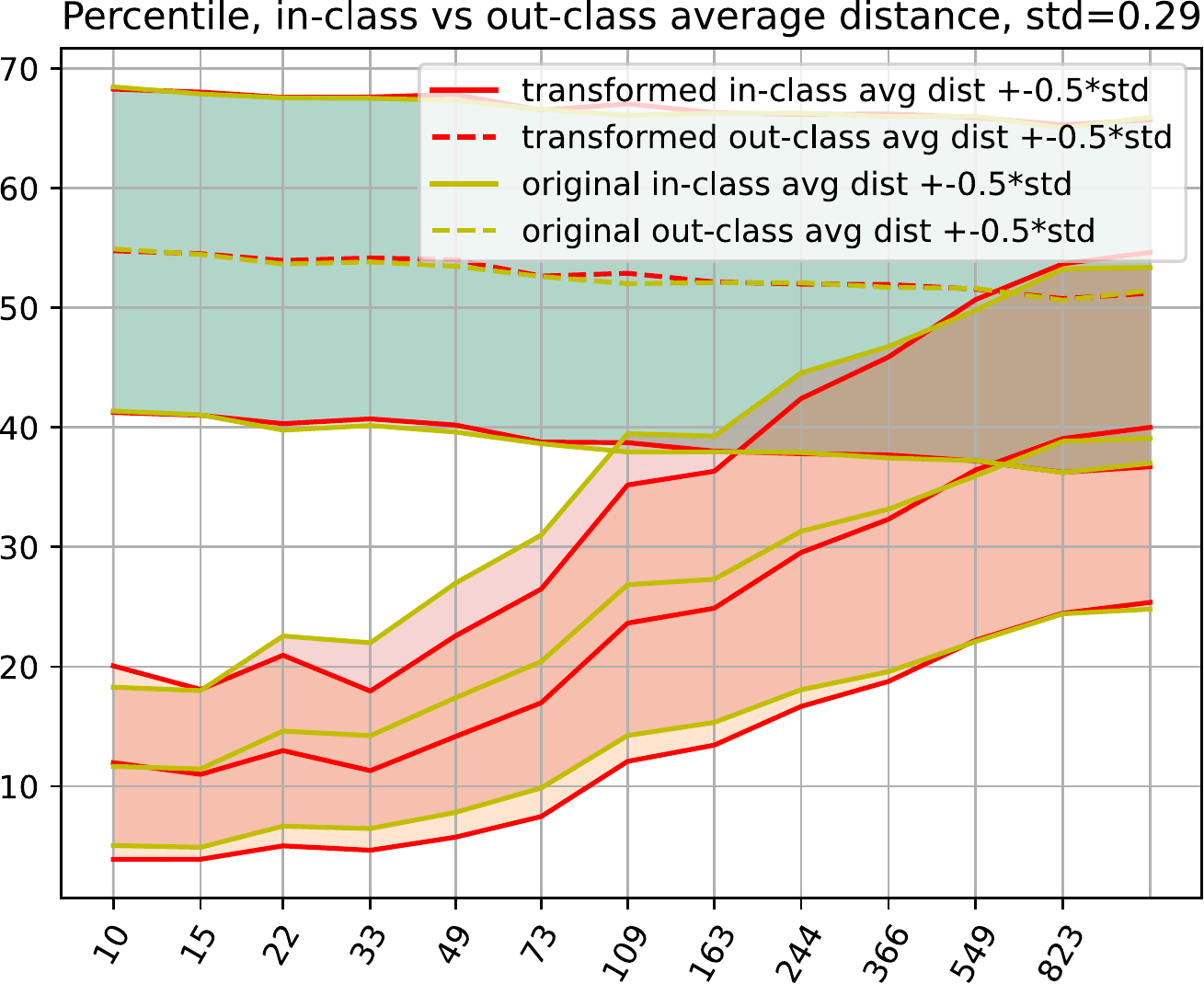} 
     ~\\ \vspace{-14pt}
\end{tabular}
   ~\\
   \vspace{-7pt}
   \caption{
\fontsize{9}{9} 
  \textbf{intra (in) vs. inter (out) class distances before and after SOT}. 
  A strong indicative property of an embedding that works on class (cluster) objects is its ability to reduce embedded intra-class (\pink{pink} shaded)
  pairwise feature distances compared to inter-class (\teal{green} shaded)
  ones. SOT (\red{red} lines)
  consistently improves this separation compared to the baseline (\brown{brown} lines)
   - leading to better downstream clustering and classification. 
   \textbf{x-axis} represents data dimensionality; 
   \textbf{y-axis} represents percentiles of pair-wise distances; The four panels present results for the noise standard deviations levels in 
   $\{0.15, 0.19, 0.23, 0.29\}$   
   }
    \label{fig.synthetic_exp_dist}\vspace{-2pt}
\end{figure}

Fig.~\ref{fig.synthetic_exp_dist} depicts the average percentile of the in-class and out-class distances computed by the original and the SOT points. Each panel presents the distributions of both types of distances, for instances of a different level of noise. We compute the mean (and plus-minus half-std) percentiles, with respect to the entire set of pair-wise distances, for a fixed level of in-class noise (increasing from top-left to bottom-right panels), for a range of data dimensionality (x-axis). 
Naturally, the overlap between in-class and between-class distances increases both with dimensionality and with in-class noise.
Nevertheless, across almost all sampled points, the situation is far better after SOT application (in red), compared to prior to SOT application (in brown). This can explain, in part, the effectiveness of using SOT in Euclidean-based downstream methods, like $k$-means and ProtoNet \cite{snell2017prototypical}.

\subsection{Evaluation on an extended set of measures}
\begin{figure}[]
 \includegraphics[width=1\columnwidth]{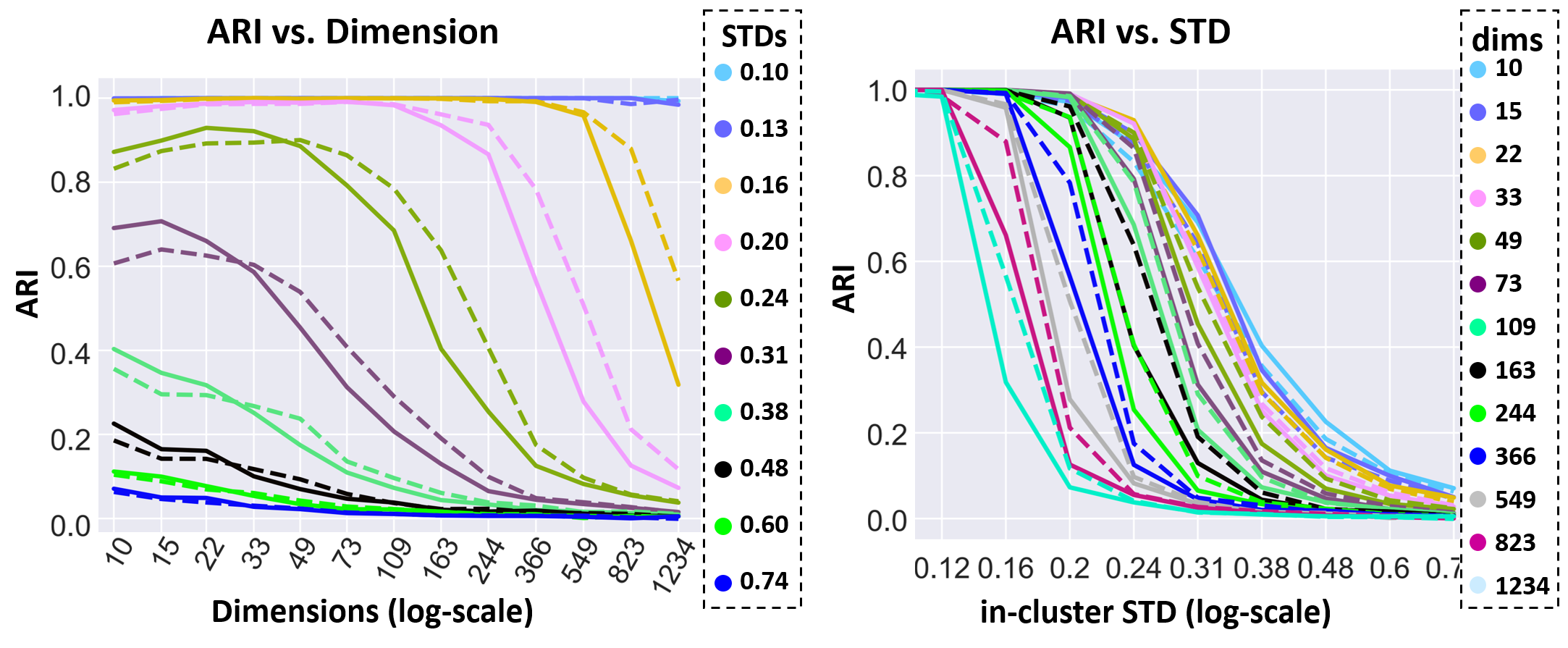} \\
 \includegraphics[width=1\columnwidth]{multi_plots.png} \\
 \includegraphics[width=1\columnwidth]{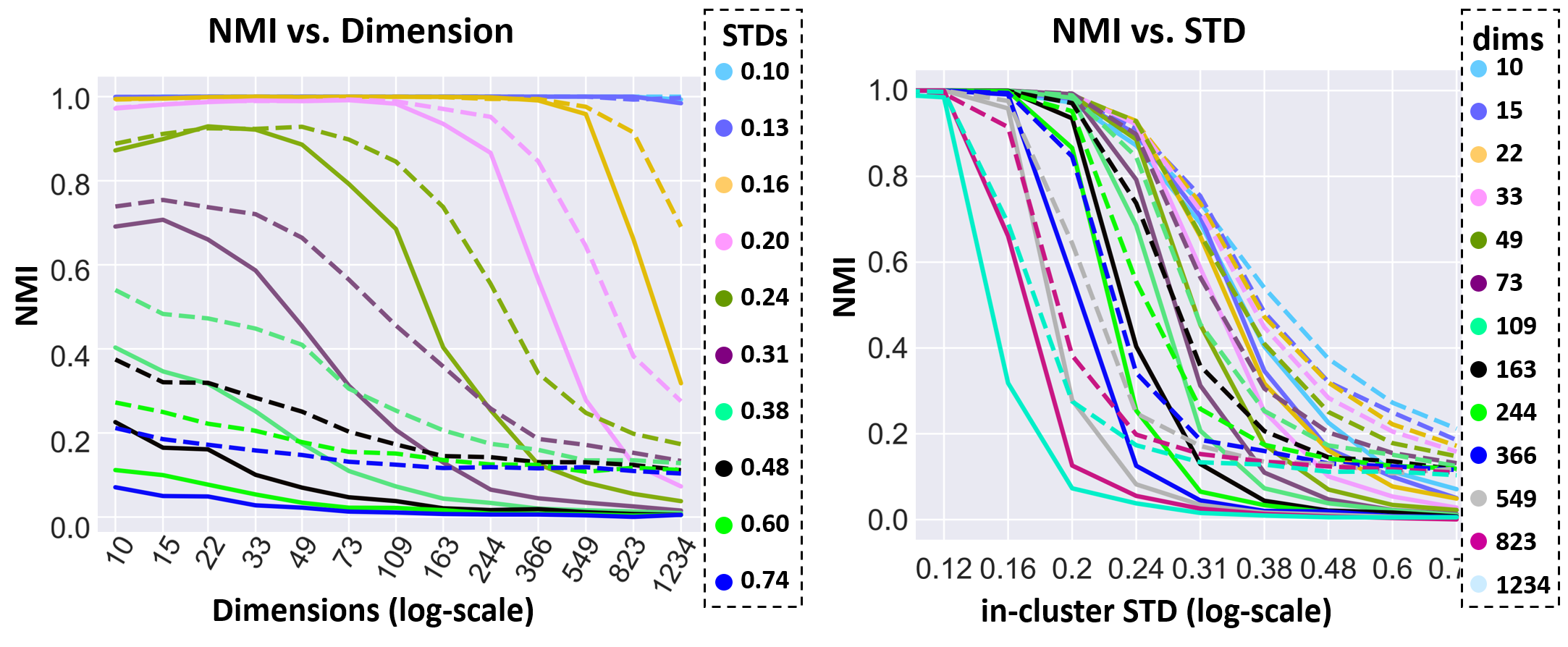} \\
 \vspace{0pt} \\\vspace{-18pt}
  \vspace{0pt}\caption{{\fontsize{9}{9} \selectfont
  \textbf{A controlled clustering experiment on the \textit{d}-dimensional sphere - Extension of results from Figure 5 of the paper, with 2 additional measures}: It can be seen that the SOT (dashed - - -) shows superior results in all aspects (see text for explanations and interpretation).
  Clustering accuracy across different noise levels $\sigma$ and dimensions $d$. $\;$\textbf{\textbf{Note}:} For each configuration, SOT is shown by a \textit{dashed} line while the baseline features are shown by a \textit{solid} line. For all 3 measures - the higher the better.
  }}
    \label{fig.synthetic_exp}\vspace{-20pt}
\end{figure}

~\\
In Fig.~\ref{fig.synthetic_exp} we evaluate the performance on additional popular clustering metrics, NMI and ARI (in addition to the accuracy measure we reported on in Figure 5 of the paper). The results shows the same trend as with accuracy, perhaps even stronger for NMI, where SOT significantly improves the clustering performance.

%% file: SOT_arxiv.bbl
\begin{thebibliography}{10}\itemsep=-1pt

\bibitem{CIFAR}
Luca Bertinetto, Joao~F. Henriques, Philip Torr, and Andrea Vedaldi.
\newblock Meta-learning with differentiable closed-form solvers.
\newblock In {\em International Conference on Learning Representations (ICLR)},
  2019.

\bibitem{caron2018deep}
Mathilde Caron, Piotr Bojanowski, Armand Joulin, and Matthijs Douze.
\newblock Deep clustering for unsupervised learning of visual features.
\newblock In {\em Proceedings of the European Conference on Computer Vision
  (ECCV)}, 2018.

\bibitem{Caron2020UnsupervisedLO}
Mathilde Caron, Ishan Misra, J. Mairal, Priya Goyal, Piotr Bojanowski, and
  Armand Joulin.
\newblock Unsupervised learning of visual features by contrasting cluster
  assignments.
\newblock {\em ArXiv}, abs/2006.09882, 2020.

\bibitem{chang2017deep}
Jianlong Chang, Lingfeng Wang, Gaofeng Meng, Shiming Xiang, and Chunhong Pan.
\newblock Deep adaptive image clustering.
\newblock In {\em Proceedings of the IEEE International Conference on Computer
  Vision (ICCV)}, 2017.

\bibitem{chen2018closer}
Wei-Yu Chen, Yen-Cheng Liu, Zsolt Kira, Yu-Chiang~Frank Wang, and Jia-Bin
  Huang.
\newblock A closer look at few-shot classification.
\newblock In {\em International Conference on Learning Representations (ICLR)},
  2018.

\bibitem{chen2021few}
Xiangyu Chen and Guanghui Wang.
\newblock Few-shot learning by integrating spatial and frequency
  representation.
\newblock {\em arXiv preprint arXiv:2105.05348}, 2021.

\bibitem{cuturi2013sinkhorn}
Marco Cuturi.
\newblock Sinkhorn distances: Lightspeed computation of optimal transport.
\newblock In {\em Advances in Neural Information Processing Systems (NeurIPS)},
  2013.

\bibitem{deng2009imagenet}
Jia Deng, Wei Dong, Richard Socher, Li-Jia Li, Kai Li, and Li Fei-Fei.
\newblock Imagenet: A large-scale hierarchical image database.
\newblock In {\em 2009 IEEE Conference on Computer Vision and Pattern
  Recognition (CVPR)}. IEEE, 2009.

\bibitem{dhillon2019baseline}
Guneet~S Dhillon, Pratik Chaudhari, Avinash Ravichandran, and Stefano Soatto.
\newblock A baseline for few-shot image classification.
\newblock In {\em International Conference on Learning Representations (ICLR)},
  2020.

\bibitem{finn2017model}
Chelsea Finn, Pieter Abbeel, and Sergey Levine.
\newblock Model-agnostic meta-learning for fast adaptation of deep networks.
\newblock In {\em International Conference on Machine Learning (ICML)}, 2017.

\bibitem{CUBSPLIT}
Yuqing Hu, Vincent Gripon, and St{\'e}phane Pateux.
\newblock Exploiting unsupervised inputs for accurate few-shot classification.
\newblock {\em ArXiv}, abs/2001.09849, 2020.

\bibitem{hu2020leveraging}
Yuqing Hu, Vincent Gripon, and St{\'e}phane Pateux.
\newblock Leveraging the feature distribution in transfer-based few-shot
  learning.
\newblock In {\em arXiv preprint arXiv:2006.03806}, 2020.

\bibitem{huang2019few}
Gabriel Huang, Hugo Larochelle, and Simon Lacoste-Julien.
\newblock Are few-shot learning benchmarks too simple? solving them without
  task supervision at test-time.
\newblock {\em arXiv preprint arXiv:1902.08605}, 2019.

\bibitem{ji2019invariant}
Xu Ji, Joao~F Henriques, and Andrea Vedaldi.
\newblock Invariant information clustering for unsupervised image
  classification and segmentation.
\newblock In {\em Proceedings of the IEEE/CVF International Conference on
  Computer Vision (ICCV)}, 2019.

\bibitem{ReNet}
Dahyun Kang, Heeseung Kwon, Juhong Min, and Minsu Cho.
\newblock Relational embedding for few-shot classification.
\newblock In {\em Proceedings of the IEEE/CVF International Conference on
  Computer Vision (ICCV)}, 2021.

\bibitem{khan2021transformers}
Salman Khan, Muzammal Naseer, Munawar Hayat, Syed~Waqas Zamir, Fahad~Shahbaz
  Khan, and Mubarak Shah.
\newblock Transformers in vision: A survey.
\newblock {\em arXiv preprint arXiv:2101.01169}, 2021.

\bibitem{kingma2014adam}
Diederik~P Kingma and Jimmy Ba.
\newblock Adam: A method for stochastic optimization.
\newblock {\em arXiv preprint arXiv:1412.6980}, 2014.

\bibitem{kolkin2019style}
Nicholas Kolkin, Jason Salavon, and Gregory Shakhnarovich.
\newblock Style transfer by relaxed optimal transport and self-similarity.
\newblock In {\em Proceedings of the IEEE/CVF Conference on Computer Vision and
  Pattern Recognition (CVPR)}, 2019.

\bibitem{korman2015coherency}
Simon Korman and Shai Avidan.
\newblock Coherency sensitive hashing.
\newblock {\em IEEE Transactions on Pattern Analysis and Machine Intelligence
  (PAMI)}, 2015.

\bibitem{krizhevsky2009learning}
Alex Krizhevsky and Geoffrey Hinton.
\newblock Learning multiple layers of features from tiny images.
\newblock 2009.

\bibitem{kuhn1955hungarian}
Harold~W Kuhn.
\newblock The hungarian method for the assignment problem.
\newblock {\em Naval Research Logistics Quarterly}, 2, 1955.

\bibitem{lee2019set}
Juho Lee, Yoonho Lee, Jungtaek Kim, Adam Kosiorek, Seungjin Choi, and Yee~Whye
  Teh.
\newblock Set transformer: A framework for attention-based
  permutation-invariant neural networks.
\newblock In {\em International Conference on Machine Learning (ICML)}, 2019.

\bibitem{CUHK03}
Wei Li, Rui Zhao, Tong Xiao, and Xiaogang Wang.
\newblock Deepreid: Deep filter pairing neural network for person
  re-identification.
\newblock In {\em Proceedings of the IEEE Conference on Computer Vision and
  Pattern Recognition (CVPR)}, 2014.

\bibitem{mangla2020charting}
Puneet Mangla, Nupur Kumari, Abhishek Sinha, Mayank Singh, Balaji
  Krishnamurthy, and Vineeth~N Balasubramanian.
\newblock Charting the right manifold: Manifold mixup for few-shot learning.
\newblock In {\em Proceedings of the IEEE/CVF Winter Conference on Applications
  of Computer Vision (WACV)}, 2020.

\bibitem{maron2020learning}
Haggai Maron, Or Litany, Gal Chechik, and Ethan Fetaya.
\newblock On learning sets of symmetric elements.
\newblock In {\em International Conference on Machine Learning (ICML)}, 2020.

\bibitem{mialon2021trainable}
Gr{\'e}goire Mialon, Dexiong Chen, Alexandre d'Aspremont, and Julien Mairal.
\newblock A trainable optimal transport embedding for feature aggregation and
  its relationship to attention.
\newblock In {\em International Conference on Learning Representations (ICLR)},
  2021.

\bibitem{qi2017pointnet}
Charles~R Qi, Hao Su, Kaichun Mo, and Leonidas~J Guibas.
\newblock Pointnet: Deep learning on point sets for 3d classification and
  segmentation.
\newblock In {\em Proceedings of the IEEE Conference on Computer Vision and
  Pattern Recognition (CVPR)}, 2017.

\bibitem{Top-DB-Net}
Rodolfo Quispe and Helio Pedrini.
\newblock Top-db-net: Top dropblock for activation enhancement in person
  re-identification.
\newblock {\em 25th International Conference on Pattern Recognition (ICPR)},
  2020.

\bibitem{ramachandran2019stand}
Prajit Ramachandran, Niki Parmar, Ashish Vaswani, Irwan Bello, Anselm Levskaya,
  and Jon Shlens.
\newblock Stand-alone self-attention in vision models.
\newblock {\em Advances in Neural Information Processing Systems (NeurIPS)},
  2019.

\bibitem{ravi2017optimization}
Sachin Ravi and Hugo Larochelle.
\newblock Optimization as a model for few-shot learning.
\newblock In {\em International Conference on Learning Representations (ICLR)},
  2017.

\bibitem{Imagenet}
Olga Russakovsky, Jia Deng, Hao Su, Jonathan Krause, Sanjeev Satheesh, Sean Ma,
  Zhiheng Huang, Andrej Karpathy, Aditya Khosla, Michael Bernstein,
  Alexander~C. Berg, and Li Fei-Fei.
\newblock {ImageNet Large Scale Visual Recognition Challenge}.
\newblock {\em International Journal of Computer Vision (IJCV)}, 2015.

\bibitem{santoro2017simple}
Adam Santoro, David Raposo, David~G Barrett, Mateusz Malinowski, Razvan
  Pascanu, Peter Battaglia, and Timothy Lillicrap.
\newblock A simple neural network module for relational reasoning.
\newblock {\em Advances in Neural Information Processing Systems (NeurIPS)},
  2017.

\bibitem{sarlin2020superglue}
Paul-Edouard Sarlin, Daniel DeTone, Tomasz Malisiewicz, and Andrew Rabinovich.
\newblock Superglue: Learning feature matching with graph neural networks.
\newblock In {\em Proceedings of the IEEE/CVF Conference on Computer Vision and
  Pattern Recognition (CVPR)}, 2020.

\bibitem{snell2017prototypical}
Jake Snell, Kevin Swersky, and Richard Zemel.
\newblock Prototypical networks for few-shot learning.
\newblock In {\em Advances in Neural Information Processing Systems (NeurIPS)},
  2017.

\bibitem{sung2018learning}
Flood Sung, Yongxin Yang, Li Zhang, Tao Xiang, Philip~HS Torr, and Timothy~M
  Hospedales.
\newblock Learning to compare: Relation network for few-shot learning.
\newblock In {\em Proceedings of the IEEE Conference on Computer Vision and
  Pattern Recognition (CVPR)}, 2018.

\bibitem{van2008visualizing}
Laurens Van~der Maaten and Geoffrey Hinton.
\newblock Visualizing data using t-sne.
\newblock {\em Journal of Machine Learning Research (JMLR)}, 9(11), 2008.

\bibitem{van2020scan}
Wouter Van~Gansbeke, Simon Vandenhende, Stamatios Georgoulis, Marc Proesmans,
  and Luc Van~Gool.
\newblock Scan: Learning to classify images without labels.
\newblock In {\em European Conference on Computer Vision (ECCV)}. Springer,
  2020.

\bibitem{vaswani2017attention}
Ashish Vaswani, Noam Shazeer, Niki Parmar, Jakob Uszkoreit, Llion Jones,
  Aidan~N Gomez, {\L}ukasz Kaiser, and Illia Polosukhin.
\newblock Attention is all you need.
\newblock In {\em Advances in Neural Information Processing Systems (NeurIPS)},
  2017.

\bibitem{vinyals2016matching}
Oriol Vinyals, Charles Blundell, Timothy Lillicrap, Koray Kavukcuoglu, and Daan
  Wierstra.
\newblock Matching networks for one shot learning.
\newblock In {\em Proceedings of the 30th International Conference on Neural
  Information Processing Systems (NeurIPS)}, 2016.

\bibitem{CUB}
Catherine Wah, Steve Branson, Peter Welinder, Pietro Perona, and Serge~J.
  Belongie.
\newblock The caltech-ucsd birds-200-2011 dataset.
\newblock 2011.

\bibitem{xie2020differentiable}
Yujia Xie, Hanjun Dai, Minshuo Chen, Bo Dai, Tuo Zhao, Hongyuan Zha, Wei Wei,
  and Tomas Pfister.
\newblock Differentiable top-k with optimal transport.
\newblock {\em Advances in Neural Information Processing Systems (NeurIPS)},
  2020.

\bibitem{ye2020few}
Han-Jia Ye, Hexiang Hu, De-Chuan Zhan, and Fei Sha.
\newblock Few-shot learning via embedding adaptation with set-to-set functions.
\newblock In {\em Proceedings of the IEEE Conference on Computer Vision and
  Pattern Recognition (CVPR)}, 2020.

\bibitem{ye2021deep}
Mang Ye, Jianbing Shen, Gaojie Lin, Tao Xiang, Ling Shao, and Steven~CH Hoi.
\newblock Deep learning for person re-identification: A survey and outlook.
\newblock {\em IEEE Transactions on Pattern Analysis and Machine Intelligence
  (PAMI)}, 2021.

\bibitem{zaheer2017deep}
Manzil Zaheer, Satwik Kottur, Siamak Ravanbakhsh, Barnabas Poczos, Russ~R
  Salakhutdinov, and Alexander~J Smola.
\newblock Deep sets.
\newblock In {\em Advances in Neural Information Processing Systems (NeurIPS)},
  2017.

\bibitem{DeepEMD}
Chi Zhang, Yujun Cai, Guosheng Lin, and Chunhua Shen.
\newblock Deepemd: Few-shot image classification with differentiable earth
  mover's distance and structured classifiers.
\newblock In {\em IEEE/CVF Conference on Computer Vision and Pattern
  Recognition (CVPR)}, June 2020.

\bibitem{zhang2021sill}
Haipeng Zhang, Zhong Cao, Ziang Yan, and Changshui Zhang.
\newblock Sill-net: Feature augmentation with separated illumination
  representation.
\newblock {\em arXiv preprint arXiv:2102.03539}, 2021.

\bibitem{Market}
Liang Zheng, Liyue Shen, Lu Tian, Shengjin Wang, Jingdong Wang, and Qi Tian.
\newblock Scalable person re-identification: A benchmark.
\newblock In {\em 2015 IEEE International Conference on Computer Vision
  (ICCV)}, 2015.

\bibitem{Re-ranking}
Zhun Zhong, Liang Zheng, Donglin Cao, and Shaozi Li.
\newblock Re-ranking person re-identification with k-reciprocal encoding.
\newblock 2017.

\bibitem{torchreid}
Kaiyang Zhou and Tao Xiang.
\newblock Torchreid: A library for deep learning person re-identification in
  pytorch.
\newblock {\em arXiv preprint arXiv:1910.10093}, 2019.

\bibitem{ziko2020laplacian}
Imtiaz~Masud Ziko, Jose Dolz, Eric Granger, and Ismail~Ben Ayed.
\newblock Laplacian regularized few-shot learning.
\newblock In {\em International Conference on Machine Learning (ICML)}, 2020.

\end{thebibliography}
